\documentclass[10pt]{article} 
\usepackage[preprint]{tmlr}

\usepackage{amsmath,amssymb,mathtools}
\usepackage{graphicx}
\usepackage{subcaption}
\usepackage{float}

\usepackage{amsmath,amsfonts,bm}

\def\eqref#1{equation~\ref{#1}}

\def\1{\bm{1}}

\DeclareMathAlphabet{\mathsfit}{\encodingdefault}{\sfdefault}{m}{sl}
\SetMathAlphabet{\mathsfit}{bold}{\encodingdefault}{\sfdefault}{bx}{n}

\usepackage{hyperref}
\usepackage{url}

\newcommand{\quotes}[1]{``#1''}

\def\month{MM}  
\def\year{YYYY} 
\def\openreview{\url{https://openreview.net/forum?id=XXXX}} 

\title{Latent States in Neural Networks: Recovering the Temporal Structure of Drifting Data from Model Weights}

\author{\name Kevin Guan \email kevinguan@princeton.edu \\
      \addr Department of Mathematics, Princeton University}

\begin{document}

\maketitle
\begin{abstract}
A temporally drifting data stream may pass through discrete regimes rather than changing continuously. We ask whether such regimes are recoverable from the weights of models trained on the stream, using a hidden Markov model (HMM) fit to the chronologically ordered trajectory of those weights. We study this question in two domains known to drift over time: multimodal misinformation detection, using the Fakeddit dataset; and sentiment analysis, using the Yelp dataset. We train classifiers on consecutive temporal windows and fit an HMM to the trajectory of their aligned weights, recovering latent states that partition each timeline into coherent phases. On both datasets, classifiers generalize better to data from windows sharing the state of their training window than to windows across state boundaries. This within-state transfer advantage survives a control for temporal proximity and modestly exceeds the advantage recovered by a naive partition into contiguous states of equal size. Although the states are estimated solely from model weights, they correlate more strongly with shifts in the data's class distribution than with the weight-space geometry used to estimate them. After class divergence and lag are residualized out, the within-state advantage exceeds its permutation null on both tasks, indicating that the states recover structure relevant to transfer beyond the data distribution. Every effect replicates on both tasks but is attenuated on Yelp, whose label distribution is more temporally stable.
\end{abstract}

\section{Introduction}
Training dynamics in learning systems are frequently characterized by phase transitions rather than continuous evolution. This same structure appears in biological learners: as humans perceive continuous experience, they spontaneously parse it into discrete events separated by boundaries. \citet{Baldassano2017} showed that this event structure is recoverable from cortical activity by fitting a hidden Markov model (HMM) that locates boundaries as transitions between latent states, without reference to the stimulus. These boundaries also have behavioral consequences: representations within an event cohere and transfer, while those across a boundary bind less readily \citep{DuBrow2013, Ezzyat2014}. 

Our study investigates whether the same event structure arises in an artificial neural network exposed to a stream of data that changes over time.\footnote{Code available at \url{https://github.com/keverage-guan/latent_states}.} If the data distribution transitions through coherent regimes, then a series of models trained on successive slices of the stream should trace a trajectory through the models' weight space that reorganizes at event boundaries. An HMM fit to that trajectory should then, by the same reasoning as \citet{Baldassano2017}, recover the regimes as latent states. We would also expect the analogous relationship between latent states and functional behavior to hold: a neural network should generalize better, i.e., degrade less, when the time periods of its training data and its test data fall within the same latent state than when they straddle a boundary. 

Earlier work fitting an HMM to a network reads out either its
activations during inference \citep{krakovna2016increasing} or scalar statistics of its weights during training \citep{Hu2023}. We instead index by the chronology of the data, where each emission is a model trained on a distinct slice of an evolving stream. To our knowledge, no prior work fits a
latent-state model to the weights of models trained on a temporally drifting
dataset. Additionally, we show that models generalize better to periods
sharing their latent state than to those across boundaries. This functional test, to our
knowledge, has not been applied to the interpretation of artificial networks using latent-state models.

We evaluate this framework in two domains in which content is known to drift temporally: misinformation detection and sentiment analysis. For each dataset, we train classifiers independently on 
consecutive windows, align and compress the
resulting weight vectors so that they are comparable across runs, and fit
an HMM to the chronologically ordered sequence. The recovered states
divide the timeline into phases within which models generalize more reliably. This advantage
withstands controls for temporal proximity and class distribution, and it holds compared against a naive equal-size
partition of the same windows.
\section{Related Work}
\label{sec:background}

\subsection{Event Segmentation}
\label{sec:event_seg}
Our approach is motivated by work in cognitive neuroscience, which finds that people spontaneously divide continuous experience into discrete events separated by boundaries. \citet{Baldassano2017} fit a hidden Markov
model to fMRI activity recorded during naturalistic narrative perception and
showed that cortical activity settles into stable patterns punctuated by abrupt
transitions, recovering these event boundaries solely from neural activity
without reference to any features of the stimulus.
What makes this segmentation consequential is that the recovered boundaries shape how experience is later retrieved. Memory of the order of events is sharper
within an event than across a boundary \citep{DuBrow2013}. \citet{Ezzyat2014} tie this to a specific neural signal, showing
that the stability of hippocampal activity patterns across a temporal gap predicts
whether the items it bridges are later bound together in memory. 
\subsection{Temporal Shift in Machine Learning}
A separate body of literature documents how machine learning models trained
on data from one period lose accuracy as the world they were fit to moves on. \citet{Won2018} develop mechanisms for anticipating temporal drift, predicting the movement of a classifier's decision boundary over time. As \citet{rottger-pierrehumbert-2021-temporal-adaptation} discuss, temporal shift has been documented across many tasks, including document classification \citep{huang-paul-2018-examining}, demographic prediction \citep{jaidka-etal-2018-diachronic}, and hate speech detection \citep{florio2020time}. Most relevant to the current work are
\citet{stepanova-ross-2023-temporal}, who study multimodal
misinformation detection, and \citet{lukes-sogaard-2018-sentiment}, who study sentiment analysis. \citet{stepanova-ross-2023-temporal} show that models evaluated on chronologically later data degrade relative to a random split, with the loss falling
disproportionately on minority classes. Similarly, \citet{lukes-sogaard-2018-sentiment} show that lexical features shift in polarity over time, degrading the performance of sentiment classifiers trained on older data.

\subsection{Phase Transitions and Latent-State Models in Learning}
The notion that a learning process passes through discrete regimes rather than evolving smoothly has substantial precedent in machine learning \citep{LiuUeda2022, Rubin2024, geiger2018jamming}. A separate line of work uses hidden Markov models and related latent-state
models to interpret a neural network's internal evolution. \citet{krakovna2016increasing} fit continuous-emission HMMs to the hidden-state sequences
of an LSTM both as a post-hoc interpretation method and
as a jointly trained hybrid, recovering discrete states that segment text into
interpretable features such as punctuation, comment symbols, and indentation. \citet{Hu2023} fit an HMM to sequences of scalar
training statistics---the norm, mean, and variance of a network's weights during
optimization---and recover latent states whose transitions mark significant
changes during training, including \quotes{detour} states that slow convergence.
\subsection{Misinformation Detection and Sentiment Analysis}
For misinformation detection, we use Fakeddit \citep{nakamura-etal-2020-fakeddit}, a multimodal benchmark of paired image-and-text Reddit posts. \citet{stepanova-ross-2023-temporal} show that classifiers trained on this data degrade when evaluated on chronologically later posts relative to a random split, with the loss concentrated in minority classes. The detection task itself remains an active area of research: \citet{Liu2024FKAOwl}, for instance, note that the wide distributional spread of machine-generated multimodal misinformation demands detectors that generalize across domains, and augment large vision-language models with forgery-specific knowledge.

For sentiment analysis, we use the Yelp review dataset \citep{yelp_open_dataset}, a 5-class task over review text. While no prior work has documented temporal drift in Yelp specifically, \citet{lukes-sogaard-2018-sentiment} establish a closely related effect in the Amazon review dataset \citep{ni-etal-2019-justifying}, showing that the lexical features driving sentiment polarity shift over time, with the effect most pronounced in sparse models. We engage with these
domains primarily as settings in which temporal drift is already established, although the framework we develop is in principle
agnostic to the particular task.
\section{Methodology}
\label{sec:methods}
\subsection{Datasets}
\label{sec:datasets}
For misinformation detection, we use Fakeddit \citep{nakamura-etal-2020-fakeddit}, a multimodal benchmark for misinformation detection assembled from a curated set of Reddit subreddits. Each post consists of a short text title and an associated image. Labels are assigned at three granularities: 2-way, 3-way, and 6-way. The present work uses the 6-way scheme, in which posts are categorized as true content, satire/parody, false connection, imposter content, manipulated content, or misleading content. For sentiment analysis, we use the Yelp review dataset \citep{yelp_open_dataset}, a text-only task consisting of user-submitted business reviews. Labels correspond to the star rating assigned by the reviewer, from one to five stars.

\subsection{Temporal Windowing}
\label{sec:temporal-windowing}
We apply a windowing procedure to each dataset, parameterized by a calendar window duration and a per-window sample size. Each dataset's samples are partitioned into non-overlapping, chronologically ordered windows of the dataset's window duration, with each sample assigned to a window by its timestamp. For each dataset, we identify the longest contiguous run of windows whose sample counts meet a minimum-count threshold (9,000 for Fakeddit, 10,000 for Yelp), and restrict analysis to that range. Windows are then downsampled to the size of the smallest window in the range via stratified subsampling on the class labels. The window duration and the minimum-count threshold were chosen separately for each dataset. Table~\ref{tab:windowing} summarizes the configuration for each dataset.

\begin{table}[ht]
\caption{Temporal windowing configuration for each dataset.}
\centering
\begin{tabular}{|l|c|c|c|c|c|}
\hline
\textbf{Dataset} & \textbf{Window} & \textbf{\# Windows} & \textbf{Samples/Window} & \textbf{Classes} & \textbf{Date Range} \\
\hline
Fakeddit & 60 days & 35 & 9,495 & 6 & Jan 2013 -- Oct 2018 \\
Yelp & 90 days & 56 & 12,801 & 5 & May 2008 -- Feb 2022 \\
\hline
\end{tabular}
\label{tab:windowing}
\end{table}

For Fakeddit, we note that small minority-class counts in early windows may introduce noise into per-window model estimates: imposter content is absent before window 14, and satire is sparse in early windows, with as few as nine samples in some windows after downsampling.
\subsection{Model Training}
\label{sec:model-training}
For each time window, a classifier is trained independently on that window's downsampled data. Each window's model is trained with 10 different seeds, so that the model weights are determined by the window's training data and a random seed. Models are optimized with cross-entropy loss and the Adam optimizer, and are trained for 20 epochs without early stopping, at a batch size of 256.

\paragraph{Fakeddit} For each of the 35 time windows, a 6-way multimodal multilayer perceptron (MLP) classifier is independently trained on concatenated RoBERTa \citep{Sanh2019DistilBERT} and ResNet-50 \citep{he2016deep} features, using the architecture of \citet{stepanova-ross-2023-temporal} (Appendix~\ref{app:replication}). The hidden width is fixed at $n=1024$, and training uses a learning rate of $10^{-4}$, the hyperparameters reported by \citet{stepanova-ross-2023-temporal} as optimal for 6-way temporal classification, so that all windows share a common capacity.

\paragraph{Yelp} For each of the 56 time windows, a 5-way MLP classifier is independently trained on RoBERTa features of the review text, consisting of a single hidden layer and a 5-way output. Prior work does not supply an architecture for the Yelp dataset, so hyperparameters were selected by grid search. A held-out pool was drawn from the dataset by stratified sampling on the class labels, with train and validation set sizes matched to the sample size in the main experiments (12,801). We searched learning rates $\mathrm{lr}\in \{10^{-2}, 10^{-3}, 10^{-4}, 10^{-5}\}$ and hidden widths $n=2^i$ for $i \in \{5, 6, 7, \dots, 14\}$. The best-performing configuration on the validation set was $\mathrm{lr}=10^{-3}$ with $n=256$, which was used for all 56 windows across the 10 training seeds.
\subsection{Weight Extraction and Processing}
\label{sec:weight_processing}
We extract the weights of each trained model and convert them into a low-dimensional, cross-comparable vector suitable for HMM fitting. Our approach is conceptually inspired by the methods of \citet{Baldassano2017}. For each window-seed pair, we extract the four weight tensors of the MLP classifier: the hidden weight matrix $W_h \in \mathbb R^{n \times d}$, hidden bias $b_h \in \mathbb R^{n}$, output weight matrix $W_o \in \mathbb R^{c \times n}$, and output bias $b_o \in \mathbb R^{c}$, where $d$ is the input feature dimension, $n$ the hidden width, and $c$ the number of classes. We then concatenate them into a flat vector of dimension $n(d+1) + c(n+1)$. For Fakeddit, $d = 2816$ for the concatenated RoBERTa and ResNet-50 features, $n = 1024$, and $c = 6$. For Yelp, $d = 768$ for the RoBERTa features, $n = 256$, and $c = 5$.

\paragraph{Permutation Alignment}
Hidden units in a feed-forward network are symmetric under permutation \citep{Entezari2022}, hence weight vectors must be aligned across seeds before dimensionality reduction. The hidden units are summarized by an alignment \quotes{signature} formed by concatenating each unit's incoming and outgoing weights.
Within each window, all seeds' models are aligned to a synthetic centroid by iterative Hungarian matching \citep{Ainsworth2023}. The recovered permutations are applied consistently across $W_h, b_h$, and $W_o$. The output biases $b_o$, which carry no hidden-unit indexing, are left unchanged. The signature construction and matching objective are given in Appendix~\ref{app:formalism-align}.

\paragraph{Dimensionality Reduction and Normalization}
Following alignment, the full set of flattened weight vectors for each dataset (35 windows $\times$ 10 seeds for Fakeddit; 56 windows $\times$ 10 seeds for Yelp) is subjected to principal component analysis (PCA). The number of components to retain was determined by inspecting the elbow of the scree plot for each dataset. The projected PCA coordinates are z-score normalized to zero mean and unit variance per component. The normalizer was fit on all aligned, projected weight vectors for the corresponding dataset.
\subsection{HMM Fitting and Model Selection}
\label{sec:hmm_fitting}
The sequence of per-window weight representations is treated as an observable emission sequence, and an HMM is used to infer the underlying latent states. We constrain the HMM to a left-to-right topology: the transition matrix satisfies $a_{ij} = 0$ unless $j \in \{i, i+1\}$, so that at each step the chain either remains in its current state or advances to the next one, and the initial-state distribution places all mass on the first state. This encodes the event-segmentation assumption of \citet{Baldassano2017}: the data stream is chronological and non-recurrent, hence a latent regime, once left, should not recur.

\paragraph{Model Selection} The number of latent states $K$ was selected using two criteria: the mean held-out log-likelihood under leave-one-seed-out cross-validation, and the Bayesian information criterion (BIC). For each candidate value of $K$---evaluated over the range from 2 to 12 for Fakeddit, and from 2 to 18 for Yelp---the HMM was trained on the concatenated per-seed weight sequences from 9 of the 10 seeds, and the per-observation log-likelihood of the held-out seed's sequence was computed. This was repeated with each seed left out, and the mean held-out log-likelihood was recorded for each $K$. Multiple random initializations were fit for each $K$ and for each fold, retaining the best-fitting solution by training log-likelihood to reduce sensitivity to initialization. We used BIC as the primary criterion and took its minimizing $K$, and then verified that this value fell at or near the elbow of the mean held-out log-likelihood curve.

\paragraph{Model Fitting} For HMM fitting, the 10 per-seed PCA vectors within each window were averaged to produce a centroid for each window, yielding a sequence of centroid vectors in the z-scored PCA space. The final HMM was fit with the Baum-Welch algorithm on this sequence of per-window centroid vectors, subject to the left-to-right constraint above, using Gaussian emissions with a diagonal covariance. The model is parameterized by initial-state probabilities, a state-transition matrix, and the per-state Gaussian means and covariances, with the zero entries of the transition matrix and the initial distribution held fixed. Multiple random initializations were fit, and the highest-likelihood solution was kept. The Viterbi algorithm was then used to decode the most probable latent-state sequence, assigning each window to a single state.
\subsection{Evaluation}
\label{sec:evaluation}
We compare classifier generalization for window pairs lying within the same HMM state versus across a state boundary. The guiding prediction is that a model should generalize better when its training window and test window belong to the same latent phase than when they straddle an event boundary. For each window $i$, the ten seed models trained on that window are evaluated on every other window $j$, recording the macro F1 of each $i \to j$ transfer per seed.

\paragraph{Column Centering} Transfer F1 conflates the intrinsic difficulty of the target window and the compatibility between source and target. We therefore column-center the transfer matrix, subtracting from each entry the mean macro F1 achieved on test window $j$ across all training windows $i \neq j$. Writing $F_{ij}$ for the raw transfer macro F1 from window $i$ to window $j$ and $W$ for the number of windows, the centered score is
\begin{equation}
\tilde{F}_{ij} \;=\; F_{ij} \;-\; \frac{1}{W-1}\sum_{ i'=1, i' \neq j}^{W} F_{i'j}.
\label{eq:centering}
\end{equation}
All downstream statistics are computed on centered F1. From this point forward, ``F1'' refers to centered F1 unless explicitly stated otherwise.

\paragraph{Within/Across Comparison} Off-diagonal entries of the matrix are partitioned into within-state pairs, in which the train and test windows share an HMM state, and across-state pairs, in which the train and test windows belong to different HMM states. The significance of the within- versus across-state performance difference is assessed by permutation testing. We stratify the state pairs by time lag and shuffle within each stratum. Each off-diagonal window pair $i \to j$, evaluated per seed, is assigned a within/across indicator designating whether windows $i$ and $j$ belong to the same state, along with a lag $d = |i - j|$. The observations are then grouped into strata by lag. Within each stratum, the within/across indicators are permuted independently. For each permutation, a test statistic is computed, and the observed statistic is compared to this null distribution to form a one-sided $p$-value. We refer to this test statistic as the \textit{distance-conditioned gap statistic}: it is a weighted average of the F1 gap between the within- and the across-group means at each lag, with each lag weighted by the harmonic mean of its within- and across-pair counts, so that lags sparsely populated in either group contribute proportionally less. The formal definition is given in Appendix \ref{app:formalism-statistic}.

\subsection{Comparison Against Equal-Size Segmentation}
\label{sec:equal-size}
We compare the HMM segmentation against a naive baseline. Both segmentations divide a dataset's $W$ windows into the number of states $K$ recovered by the HMM, not including unvisited states. The baseline segmentation partitions the windows into contiguous blocks of approximately equal size: each block receives $\lfloor W/K \rfloor$ windows, and the $W \bmod K$ remainder windows are allocated to the blocks from left to right. The segmentations are compared with a permutation test on the difference $D$ of their pooled gaps; the pooled gap of a segmentation is the difference between the mean within-group centered F1 and the mean across-group centered F1 over all off-diagonal pairs, which we denote $\Pi$. The aggregation weights, the statistics $\Pi$ and $D$, and the construction of the null distribution for $D$ are detailed formally in Appendix~\ref{app:formalism-equalsize}.

\subsection{Correlation Analyses}
\label{sec:correlation}
To interpret the information captured by the latent states, two dissimilarity measures were correlated with cross-window generalization, aggregated at the level of ordered state pairs. For each ordered pair of states, the mean cross-window centered macro F1 over the constituent pairs is related to (1) class distribution, measured by the Jensen-Shannon divergence (JSD) between the class distributions of the two states, and (2) weight-space, measured by the Euclidean distance between the states' PCA centroids in the z-scored PCA space. Associations were quantified with Spearman's rank correlation. Permutation $p$-values are obtained similarly to the analysis above: the dissimilarity values across state pairs were repeatedly permuted while holding the corresponding performance values fixed, and Spearman's rank correlation was recomputed, yielding a null distribution against which the observed association was compared.

\subsection{Class Distribution Stability Within and Across States}
\label{sec:stability}
The correlation analysis above operates at the level of aggregated state pairs. As a more fine-grained analysis of whether the recovered states correspond to stability in the data distribution, we examine the pairwise class distribution dissimilarity between individual windows, grouped by whether they share a state. For each unordered pair of windows $(i, j)$, we compute the Jensen-Shannon divergence between their class distributions, $\mathrm{JSD}(c_i \,\Vert\, c_j)$, and assign the pair to the within-group if windows $i$ and $j$ share an HMM state, $s_i = s_j$, and to the across-group otherwise. We compare the two groups on the mean and variance of their pairwise JSD, and assess the difference in group mean divergence, $\Delta_{\mathrm{JSD}} = \bar{D}^{a} - \bar{D}^{w}$, where $\bar{D}^{w}$ and $\bar{D}^{a}$ are the mean pairwise JSD within and across states. Significance is assessed with a label-shuffle permutation test that holds the JSD values fixed and permutes the within/across labels across window pairs. The formal group-mean definitions and permutation procedure are detailed in Appendix~\ref{app:formalism-stability}.

\subsection{Controlling for Class Distribution}
\label{sec:partial-jsd}
The analyses of Section~\ref{sec:results-corr} and Section~\ref{sec:results-jsd-stability} establish that state membership is tied to the class distribution. This raises the possibility that the within/across indicator is merely a proxy for class divergence, in which case the within-state advantage should vanish when divergence is held fixed. We test whether the within-state indicator continues to predict transfer between window pairs when controlling for class divergence and temporal lag. We use the per-pair divergence $\mathrm{JSD}(c_i \, \Vert\, c_j)$ introduced in Section~\ref{sec:stability}. The aggregated, state-level measure used in Section~\ref{sec:correlation} is unsuitable for this control, as it is identically zero for every within-state pair. The per-window measure is continuous and varies both within and across states.

\paragraph{Combined Model} For each off-diagonal window pair $i \to j$, we record its centered transfer macro F1 $\tilde{F}_{ij}$, the per-pair divergence $\mathrm{JSD}(c_i \,\Vert\, c_j)$, the lag $d = |i-j|$, and the indicator $\mathbf{1}_{\{s_i = s_j\}}$. We fit the linear model
\begin{equation}
\tilde{F}_{ij} = \beta_0 + \beta_{\mathrm{jsd}}\,\mathrm{JSD}(c_i \,\Vert\, c_j)
       + \beta_{\mathrm{lag}}\, d
       + \beta_{\mathrm{state}}\, \mathbf{1}_{\{s_i = s_j\}} + \varepsilon_{ij},
\end{equation}
and interpret $\beta_{\mathrm{state}}$ as the within-state advantage in centered macro F1 units with divergence and lag held fixed. Its significance is assessed with a Freedman-Lane partial permutation test \citep{FreedmanLane1983}, which extends the distance-conditioned shuffle in Section~\ref{sec:evaluation} to this partial-correlation setting. The permutation procedure is detailed in Appendix~\ref{app:formalism-partial}. 

\paragraph{Two-Stage Residualization} As a complementary form of the same test, we residualize centered transfer F1 on divergence and lag and apply the harmonic-weighted, lag-stratified distance-conditioned gap statistic of Section~\ref{sec:evaluation} to the residuals. The residualization is detailed in Appendix~\ref{app:formalism-partial}.
\section{Results}
\label{sec:results}
\subsection{Weight-Space Representation}
\label{sec:weight-space}
We extracted and aligned the head weights from all trained models and
applied PCA to the aligned vectors as described in
Section~\ref{sec:weight_processing}. For the Fakeddit dataset, with 350 total models, three principal components were retained at
the elbow of the scree curve (PC1: 7.35\%, PC2: 2.66\%, PC3: 1.45\%),
cumulatively capturing 11.47\% of the total variance. For the Yelp dataset, with 560 total models, three principal components were retained (PC1: 7.74\%, PC2: 2.07\%, PC3: 1.29\%),
cumulatively capturing 11.10\% of the total variance. The scree plots for both
datasets are shown in Appendix~\ref{app:supp-scree}.
Figure~\ref{fig:pca_space} shows PCA projections of the aligned weight vectors
for both datasets. Before alignment, points cluster by seed as a consequence of
permutation symmetry in the hidden layer (Figure~\ref{fig:pca_space_before},
Appendix~\ref{app:supp-prealign}). After alignment, the dominant source
of clustering shifts to the temporal window, and the points trace
a coherent trajectory through PC space.
\begin{figure}[ht]
\centering
\begin{subfigure}[t]{0.49\textwidth}
    \centering
    \includegraphics[width=\linewidth]{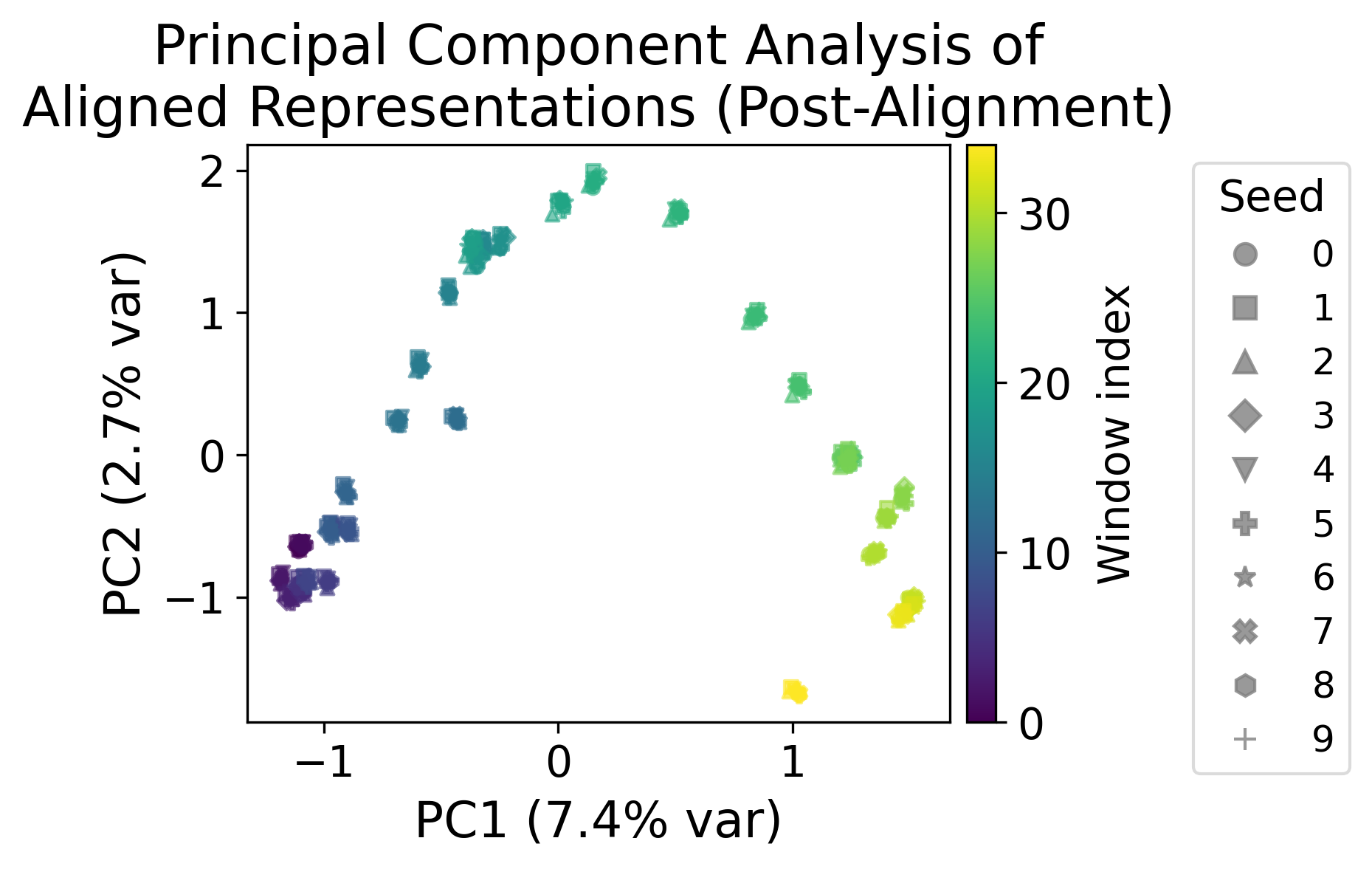}
    \caption{Fakeddit (after alignment).}
    \label{fig:pca_space_fakeddit}
\end{subfigure}
\hfill
\begin{subfigure}[t]{0.49\textwidth}
    \centering
    \includegraphics[width=\linewidth]{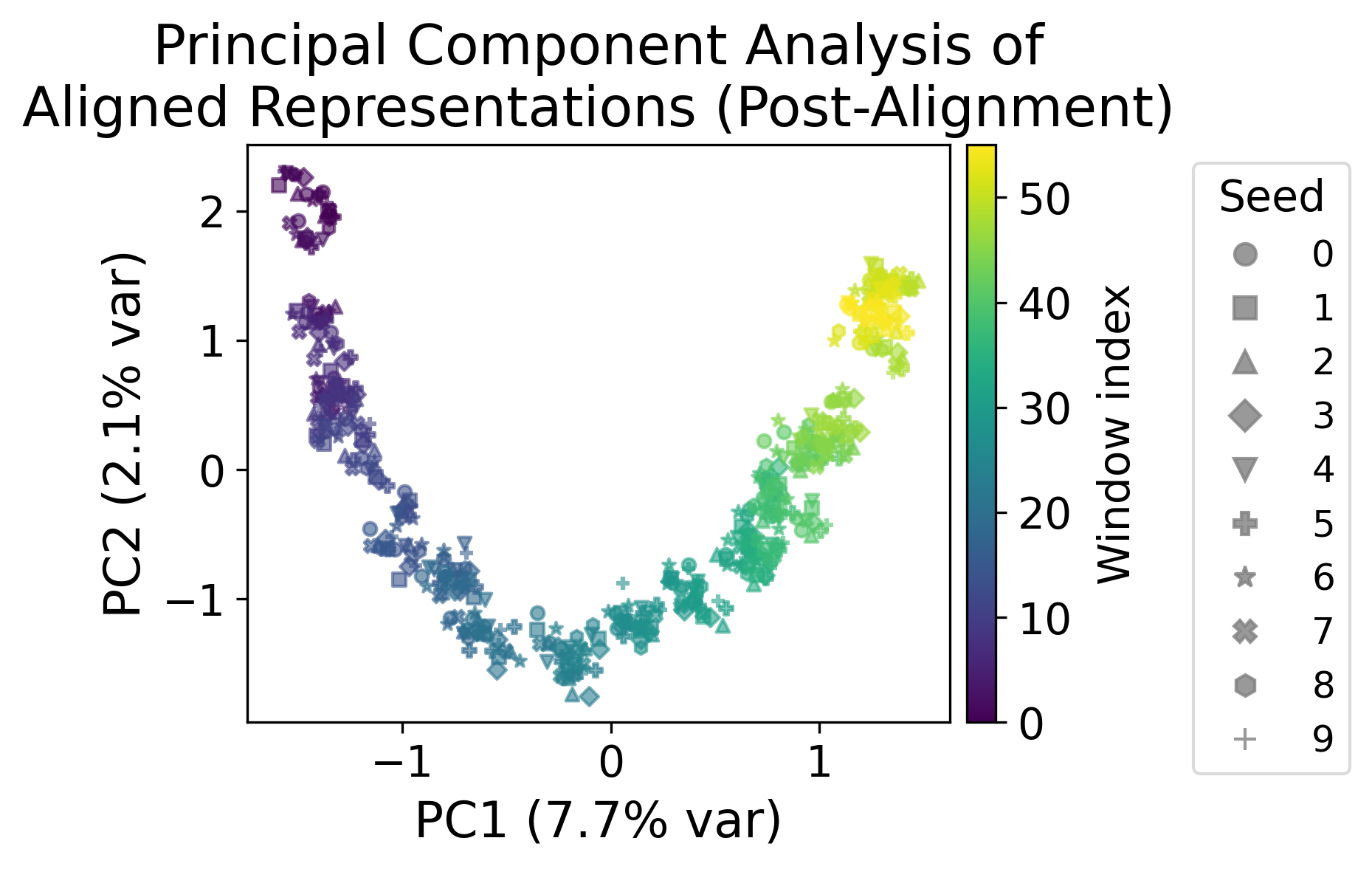}
    \caption{Yelp (after alignment).}
    \label{fig:pca_space_yelp}
\end{subfigure}
\caption{PCA projections of aligned MLP head weights (PC1 vs.\ PC2) for
(a) Fakeddit and (b) Yelp. After alignment, clustering is organized by
window, revealing coherent temporal drift in both datasets.}
\label{fig:pca_space}
\end{figure}
\subsection{Latent State Discovery}
\begin{figure}[ht]
\centering
\begin{subfigure}{0.49\textwidth}
    \centering
    \includegraphics[width=\linewidth]{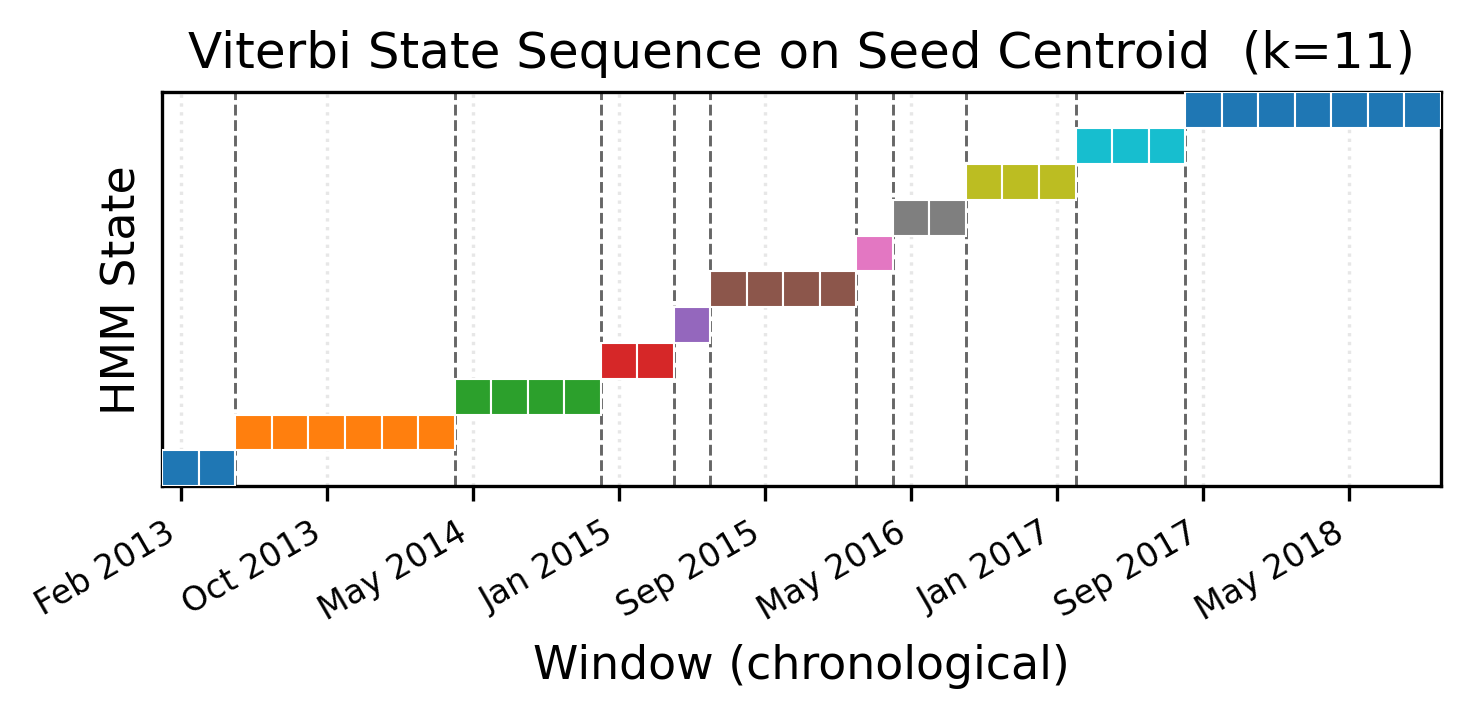}
    \caption{Fakeddit ($K=11$), 35 windows.}
    \label{fig:state_timeline}
\end{subfigure}
\hfill
\begin{subfigure}{0.49\textwidth}
    \centering
    \includegraphics[width=\linewidth]{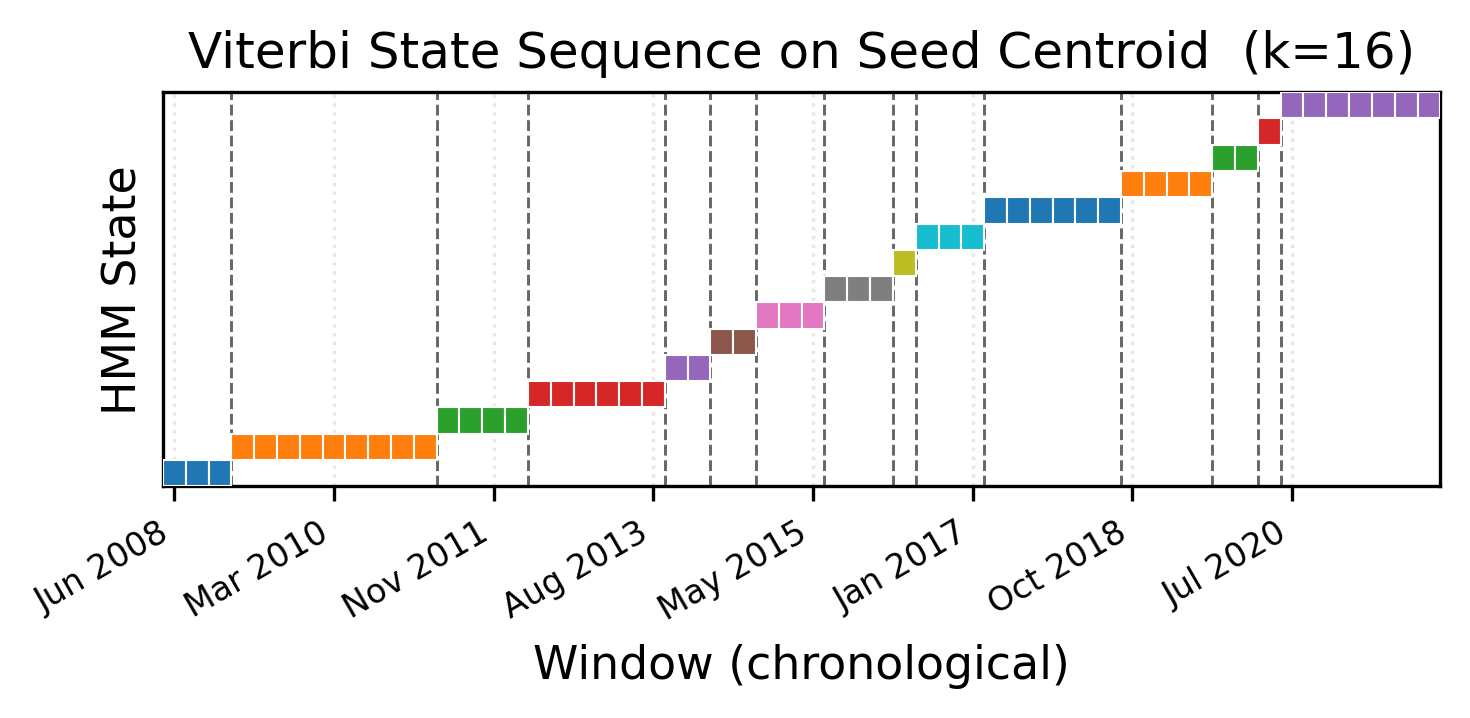}
    \caption{Yelp ($K=16$), 56 windows. State 15 is unvisited.}
    \label{fig:state_timeline_yelp}
\end{subfigure}
\caption{Viterbi-decoded HMM state sequences; color indicates the assigned state.}
\label{fig:state_timelines}
\end{figure}

An HMM was fit to the sequence of per-window
weight centroids in the z-scored PCA space. For Fakeddit, the number of states
was searched over $K \in \{2,\ldots,12\}$; BIC was minimized at $K=11$, and the leave-one-seed-out log-likelihood per observation plateaued around $K=10$--$11$, so $K=11$ was
selected (Figure~\ref{fig:hmm_select_fakeddit}, Appendix~\ref{app:supp-selection}). For Yelp, $K$ was searched
over $K \in \{2,\ldots,18\}$; BIC was minimized at $K=16$, with the log-likelihood per observation climbing marginally past that point, so $K=16$ was selected (Figure~\ref{fig:hmm_select_yelp}, Appendix~\ref{app:supp-selection}). The Viterbi algorithm was
then used to decode the most probable state sequence (Figures~\ref{fig:state_timeline}
and~\ref{fig:state_timeline_yelp}). Under the left-to-right decode, the
Yelp sequence visits only 15 of the 16 states, as state 15 is never occupied. Subsequent analyses therefore effectively concern these 15
visited states.
\subsection{Within- vs. Across-State Generalization}
\label{sec:results-wa}
\begin{figure}[ht]
\centering
\begin{subfigure}[t]{0.49\textwidth}
    \centering
    \includegraphics[width=\linewidth]{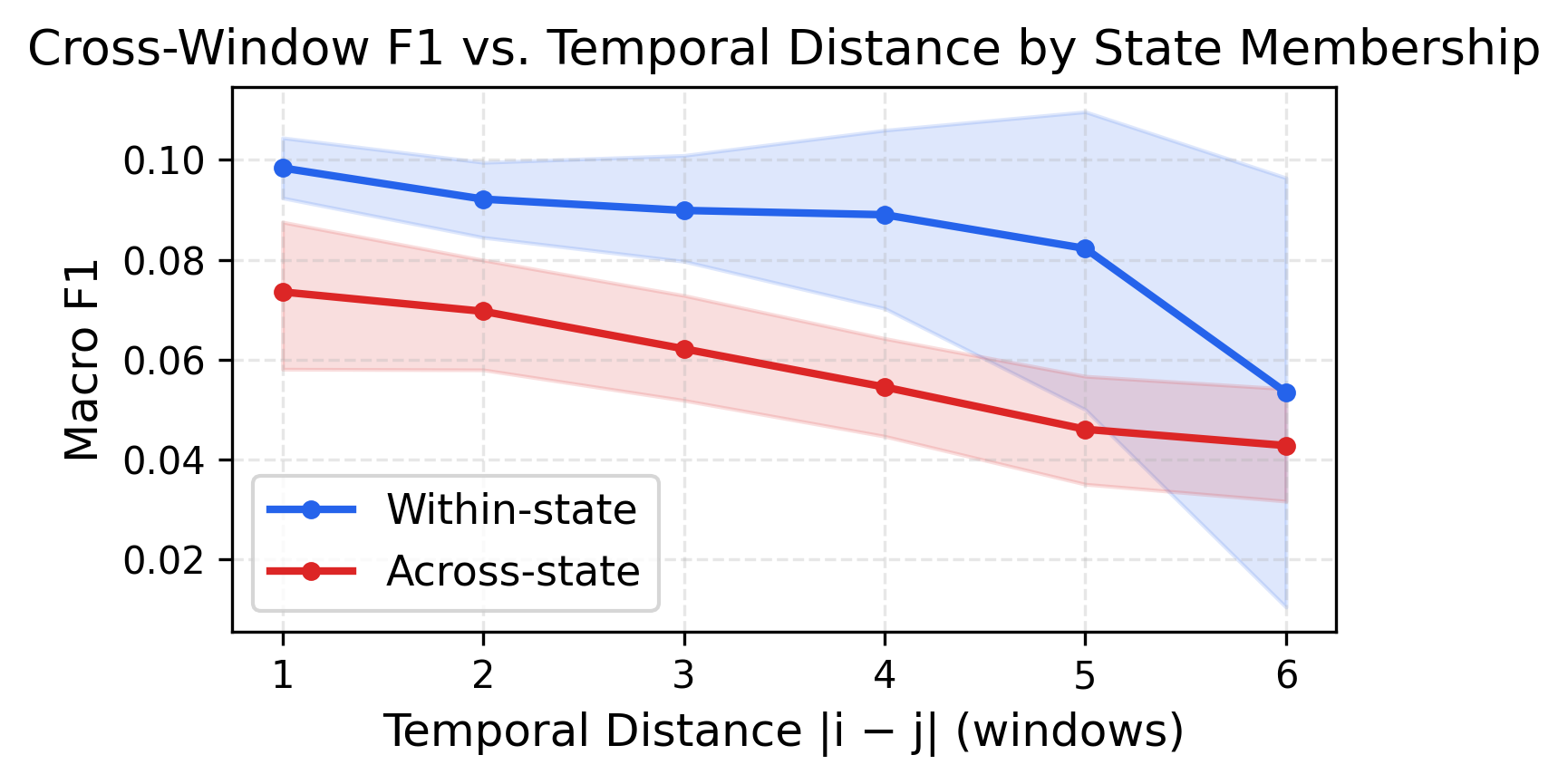}
    \caption{Fakeddit.}
    \label{fig:f1_vs_distance_fakeddit}
\end{subfigure}
\hfill
\begin{subfigure}[t]{0.49\textwidth}
    \centering
    \includegraphics[width=\linewidth]{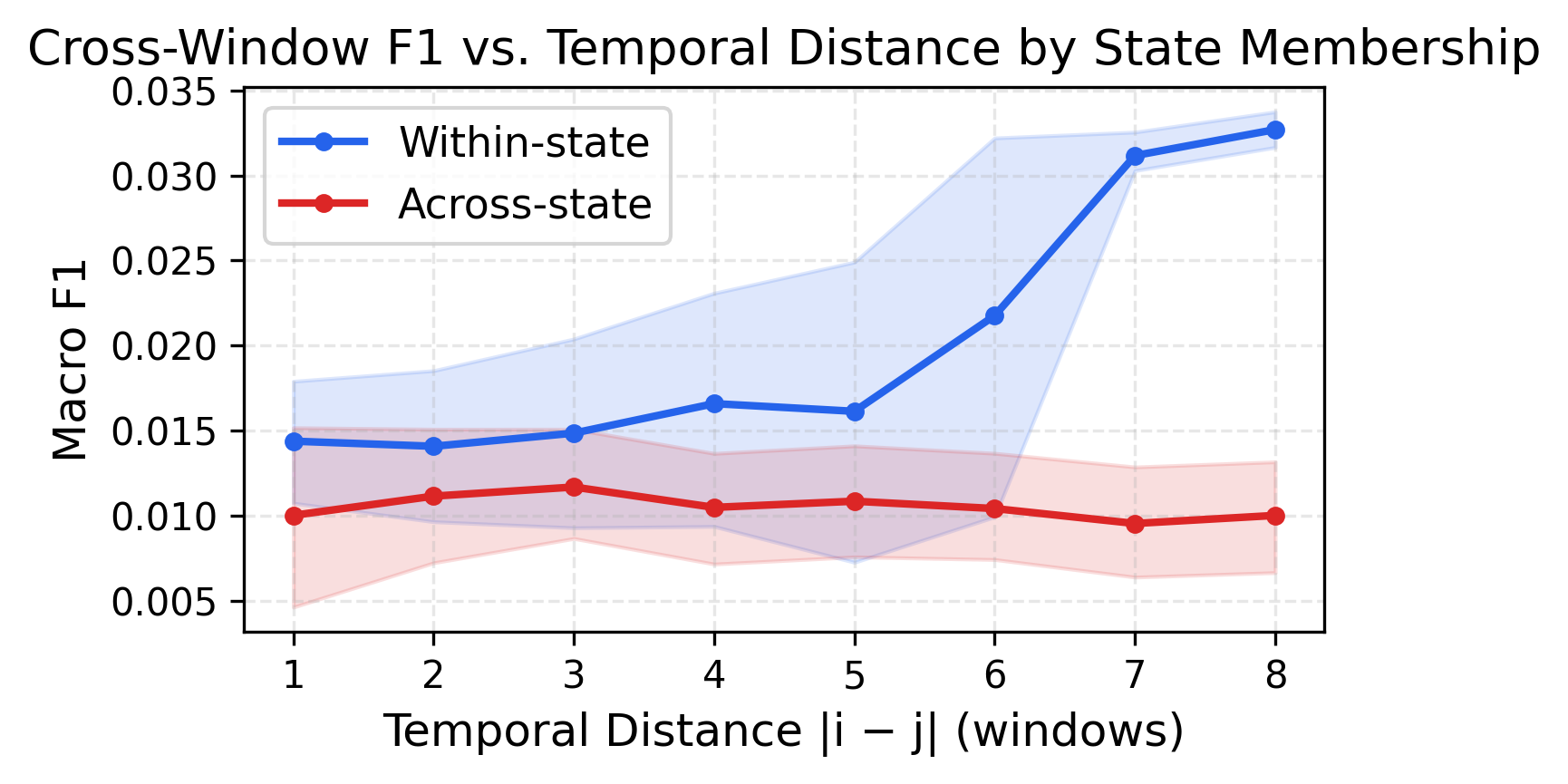}
    \caption{Yelp.}
    \label{fig:f1_vs_distance_yelp}
\end{subfigure}
\caption{Mean column-centered macro F1 for within- versus across-state
window pairs as a function of temporal lag (windows between training and
test windows). Error bars are 95\% confidence intervals.}
\label{fig:f1_vs_distance}
\end{figure}

To evaluate whether models generalize better within states than across boundaries, we computed the macro F1 of every window-to-window
transfer and partitioned pairs by whether their windows share the same state. Pooled across all off-diagonal pairs without lag weighting, Fakeddit within-state transfers achieved a mean macro F1 of $+0.0929$, compared to $-0.0162$ for across-state transfers (114 within,
1,076 across pairs), a gap of $0.1092$. For Yelp, within-state
transfers achieved $+0.0170$ against $-0.0014$ for across-state transfers
(228 within, 2,852 across pairs), a gap of $0.0184$. Figure~\ref{fig:f1_vs_distance} shows the within- and
across-state means by temporal lag.
To test the possibility that this difference is an artifact of
temporal proximity, we applied a distance-conditioned label-shuffle test (Figure~\ref{fig:perm_null},
Appendix~\ref{app:null_distributions}).
Within each temporal-lag stratum, the within/across indicators were
permuted independently while holding the
per-lag F1 distributions fixed. The distance-conditioned gap statistic
was $+0.0270$ for Fakeddit ($p < 0.0001$) and $+0.0056$ for Yelp
($p = 0.0001$).
\subsection{Comparison Against Equal-Size Segmentation}
\label{sec:results-equal}
For each segmentation we compute the pooled F1 gap between the within- and across-groups over off-diagonal pairs, and assess the difference $D$ between the pooled gaps with a direct permutation test (Figure~\ref{fig:equal_perm}, Appendix~\ref{app:null_distributions}). Because only 15 of the 16 decoded states were actually visited for Yelp, the equal-size baseline consists of 15 states. For Fakeddit, the HMM segmentation produced a pooled gap of $0.1092$, while the equal-size segmentation produced a pooled gap of $0.1041$; the difference was significant ($D = +0.0051$, $p < 0.0001$). For Yelp, the HMM segmentation produced a pooled gap of $0.0184$ against $0.0167$ for the equal-size segmentation, and the difference was again significant ($D = +0.0016$, $p < 0.0001$). The within- and across-group F1 are shown as a function of temporal lag in Figures~\ref{fig:f1_vs_distance_both_fakeddit} and~\ref{fig:f1_vs_distance_both_yelp} (Appendix~\ref{app:supp_figs}). 
\subsection{Correlation Analyses}
\label{sec:results-corr}
To interpret the dimension of dissimilarity captured by HMM state membership, we correlate cross-window transfer performance with two complementary measures of state-pair dissimilarity (Section~\ref{sec:correlation}): a data-side measure, the Jensen-Shannon divergence between the two states' class distributions; and a model-side measure, the Euclidean distance between the states' centroids in the z-scored PCA space. For Fakeddit, we aggregate over the 110 ordered off-diagonal state pairs ($11 \times 10$); for Yelp, over the 210 ordered off-diagonal state pairs ($15 \times 14$). In each case $i\!\to\!j$ and $j\!\to\!i$ are treated separately because transfer F1 is asymmetric.

\begin{figure}[ht]
\centering
\begin{subfigure}[t]{0.48\textwidth}
    \centering
    \includegraphics[width=\linewidth]{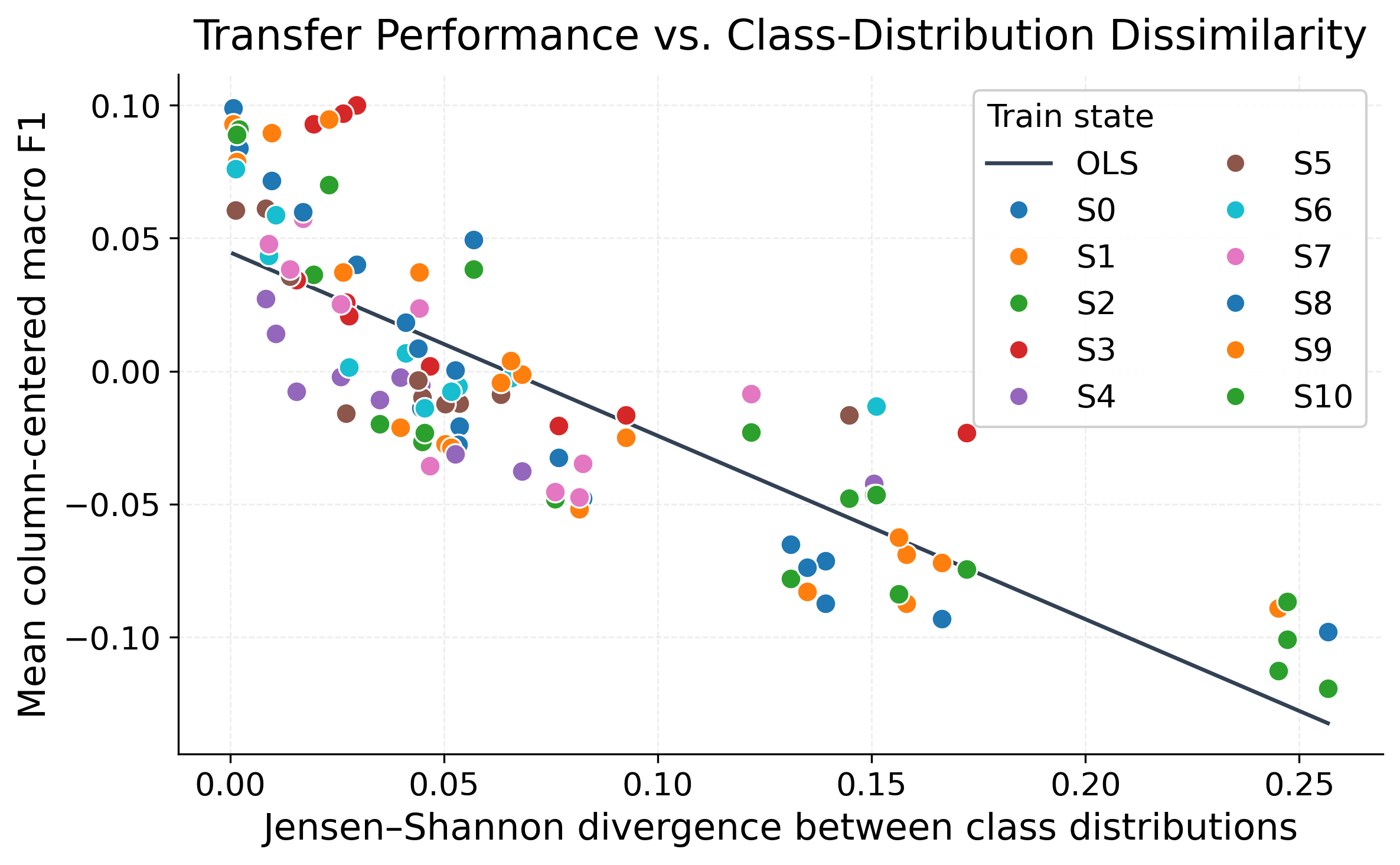}
    \caption{Class distribution divergence (JSD) vs.\ mean cross-window
    macro F1 (Spearman $\rho = -0.878$, $p < 0.0001$).}
    \label{fig:corr_jsd}
\end{subfigure}
\hfill
\begin{subfigure}[t]{0.48\textwidth}
    \centering
    \includegraphics[width=\linewidth]{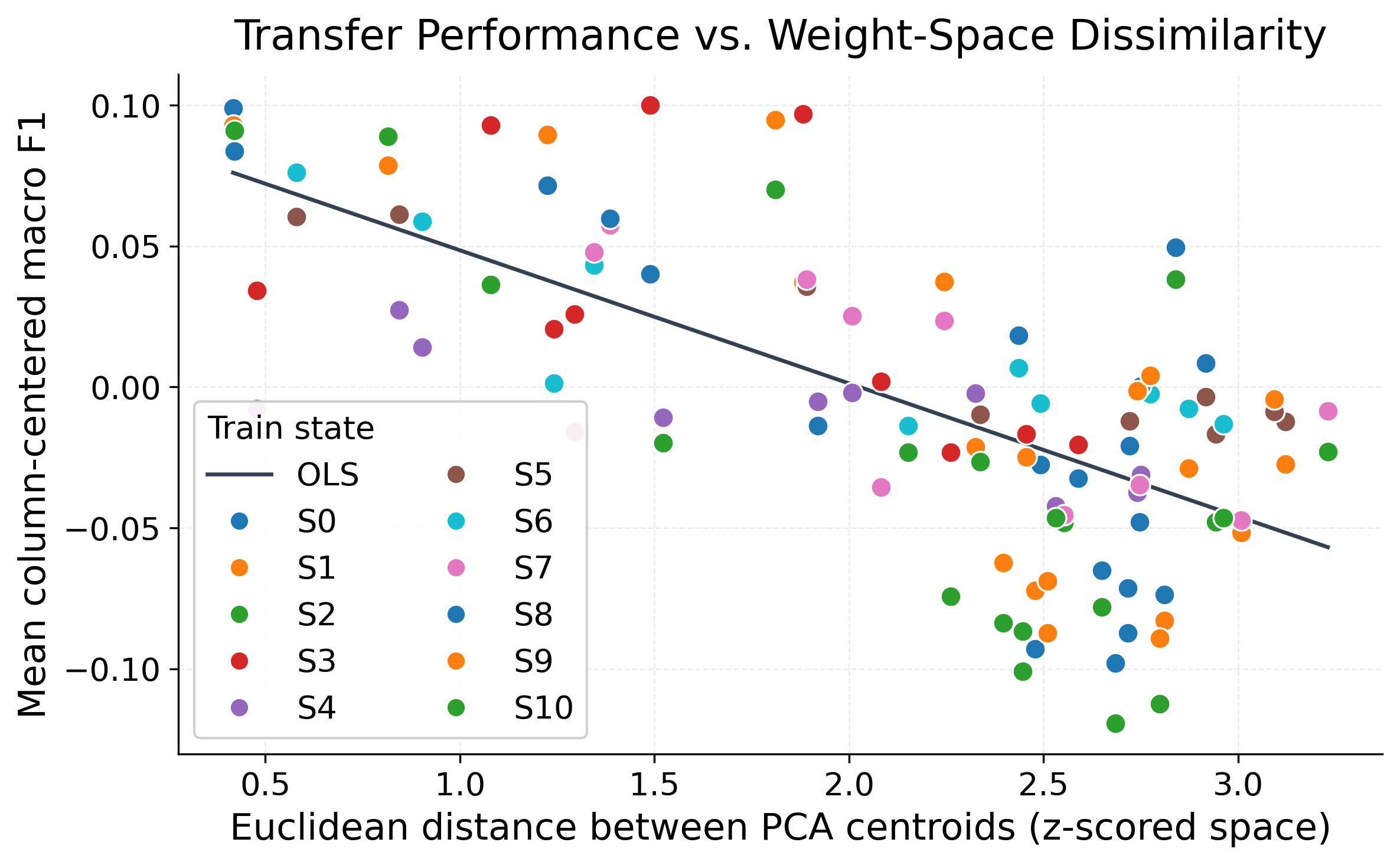}
    \caption{Weight-space centroid distance vs.\ mean cross-window
    macro F1 (Spearman $\rho = -0.617$, $p < 0.0001$).}
    \label{fig:corr_pca}
\end{subfigure}
\caption{Fakeddit: dissimilarity between state pairs versus mean cross-window
macro F1. The data-side measure (left) is a stronger predictor of transfer performance than the weight-space measure (right).}
\label{fig:corr_both_fakeddit}
\end{figure}

\begin{figure}[ht]
\centering
\begin{subfigure}[t]{0.48\textwidth}
    \centering
    \includegraphics[width=\linewidth]{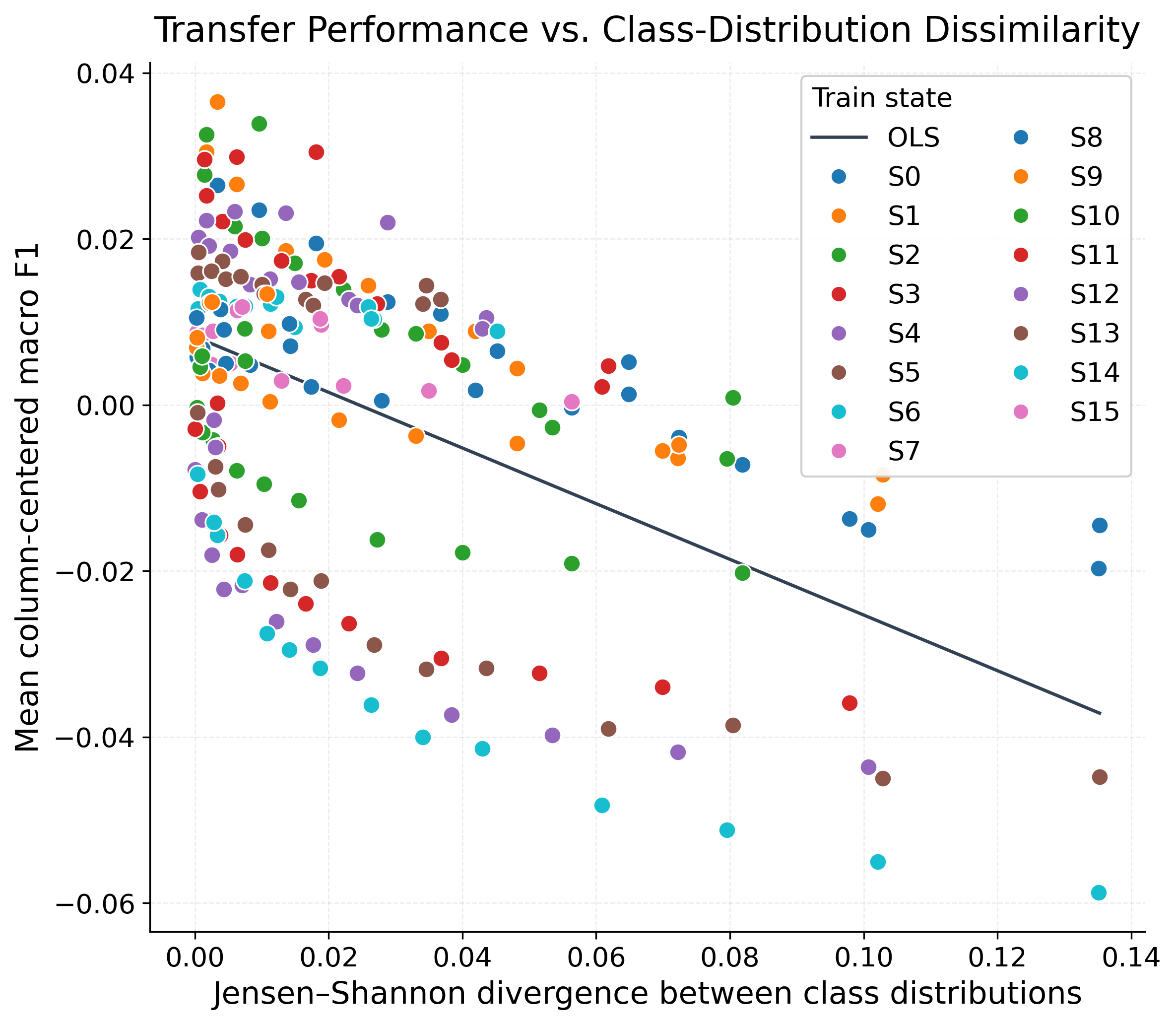}
    \caption{Class distribution divergence (JSD) vs.\ mean cross-window
    macro F1 (Spearman $\rho = -0.407$, $p < 0.0001$).}
    \label{fig:corr_jsd_yelp}
\end{subfigure}
\hfill
\begin{subfigure}[t]{0.48\textwidth}
    \centering
    \includegraphics[width=\linewidth]{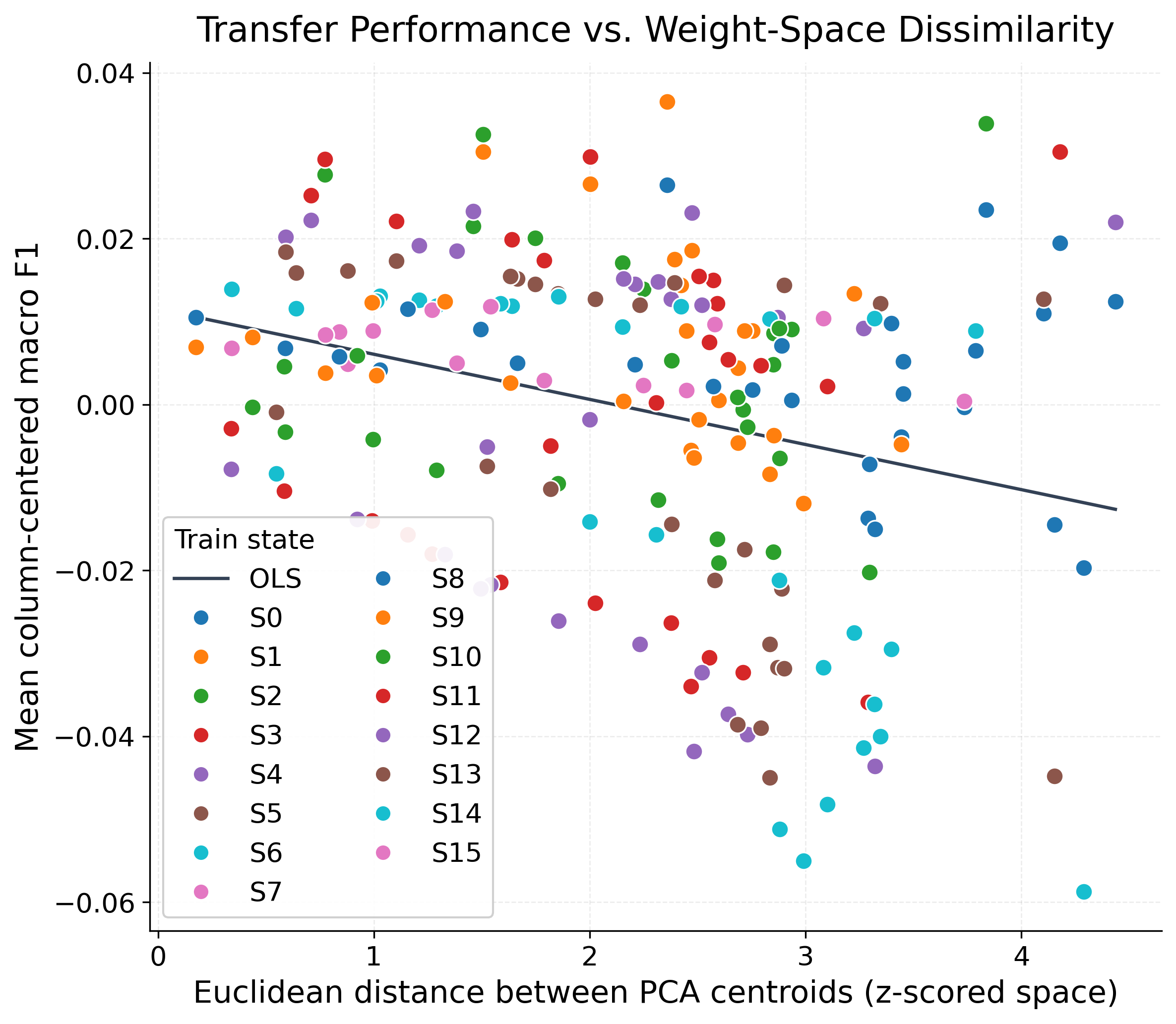}
    \caption{Weight-space centroid distance vs.\ mean cross-window
    macro F1 (Spearman $\rho = -0.285$, $p < 0.0001$).}
    \label{fig:corr_pca_yelp}
\end{subfigure}
\caption{Yelp: dissimilarity between state pairs versus mean cross-window
macro F1. Both associations run in the same direction as Fakeddit but are
substantially weaker.}
\label{fig:corr_both_yelp}
\end{figure}

Both measures correlate negatively with mean cross-window macro F1 on both datasets. For Fakeddit, the class distribution divergence was strongly and significantly associated with transfer performance (Spearman $\rho = -0.878$, permutation $p < 0.0001$; Figure~\ref{fig:corr_both_fakeddit}), as was the weight-space centroid distance ($\rho = -0.617$, permutation $p < 0.0001$). For Yelp, both associations were in the same direction and remained significant, though substantially weaker: class distribution divergence gave $\rho = -0.407$ ($p < 0.0001$; Figure~\ref{fig:corr_both_yelp}) and weight-space centroid distance gave $\rho = -0.285$ ($p < 0.0001$). Permutation null distributions for all correlations are provided in Appendix~\ref{app:null_distributions} (Figures~\ref{fig:corr_analysis_fakeddit} and~\ref{fig:corr_analysis_yelp}). On both datasets, the data-side measure was the stronger predictor of generalization, despite the HMM states having been fit entirely in weight space.
\subsection{Within-State Class Distribution Stability}
\label{sec:results-jsd-stability}
The correlation analysis shows that more dissimilar states transfer less reliably, but
does not establish how states correspond to the data distribution. To test this, we compared the pairwise
class distribution divergence of same-state window pairs against pairs
that span a boundary. For Fakeddit, within-state window pairs ($n=57$) had a mean
pairwise JSD of $0.0119$, whereas across-state pairs ($n=538$) had a mean of
$0.1053$. The two groups also differ in dispersion: the variance of within-state JSD ($3.3\times10^{-4}$) is
much smaller than that of across-state JSD
($7.9\times10^{-3}$), indicating that within-state pairs are not only closer but also far more tightly concentrated. The difference in mean pairwise JSD ($+0.0935$) lies far outside the null
distribution ($p < 0.0001$; Figure~\ref{fig:jsd_null}, Appendix~\ref{app:null_distributions}). The Yelp dataset shows the same pattern. Within-state window pairs ($n=114$) had a mean
pairwise JSD of $0.0004$ against $0.0316$ for across-state pairs ($n=1,426$), and
the variance of within-state JSD ($2.4\times10^{-7}$) was again far smaller than that of
across-state JSD ($1.1\times10^{-3}$). The difference in mean pairwise JSD ($+0.0312$)
lies far outside the null distribution ($p < 0.0001$). 
\subsection{State Membership Beyond Class Distribution}
\label{sec:results-partial}
Sections~\ref{sec:results-corr} and~\ref{sec:results-jsd-stability} establish that the recovered states approximate regimes of stable class distribution. This
invites the concern that the within-state generalization advantage of
Section~\ref{sec:results-wa} merely restates that fact: if the within/across indicator is only a proxy for class divergence, the advantage should vanish once divergence is held fixed. We tested this by
partialling per-pair class divergence and temporal lag out of transfer F1 and
asking whether state membership retains predictive power
(Section~\ref{sec:partial-jsd}). The control holds on both datasets, though the effect is much larger on Fakeddit than on Yelp.

For Fakeddit, over the ordered off-diagonal window pairs, the within-state indicator remained a strong
predictor of transfer after divergence and lag were held fixed, with a coefficient of
$\beta_{\mathrm{state}} = +0.0375$ (Freedman-Lane one-sided $p < 0.0001$; Figure~\ref{fig:combined_null}, Appendix~\ref{app:null_distributions}).
Class divergence was, as expected, a large
negative predictor ($\beta_{\mathrm{jsd}} = -0.2870$), while lag was small
and negative ($\beta_{\mathrm{lag}} = -0.0041$), indicating that the
temporal-proximity benefit on transfer is itself largely mediated by class
divergence. The two-stage
residualization agreed: the gap between the within-group and the across-group on F1 residualized for divergence and lag was $+0.0272$ under the
lag-stratified harmonic statistic ($p < 0.0001$; Figure~\ref{fig:partial_resid_null}, Appendix~\ref{app:null_distributions}).

For Yelp, the effect is present but weaker. Under the two-stage
residualization, the gap between the within- and across-groups on transfer F1
residualized for divergence and lag was $+0.0054$ ($p = 0.0001$). In the
combined model, the within-state indicator remained predictive (Figure~\ref{fig:combined_null}, Appendix~\ref{app:null_distributions}). Because divergence and lag are partly collinear
with state membership, they absorb most of the within-state signal, and the Freedman-Lane null for $\beta_{\mathrm{state}}$ is thus
centered below zero under this residualization. The observed coefficient ($\beta_{\mathrm{state}}=-0.0003$)
falls firmly in its upper tail (one-sided $p = 0.0234$), hence state membership retains a small but significant association with transfer despite a slightly negative raw coefficient. Class divergence was
again a negative predictor ($\beta_{\mathrm{jsd}} = -0.0773$) and lag was
negligible ($\beta_{\mathrm{lag}} = -0.0009$).
\section{Discussion}
\label{sec:discussion}
Our study tested whether the event-segmentation framework that describes human perception of naturalistic experience can be transposed to an artificial learner exposed to a temporally drifting data stream. Our results establish three key findings. First, after controlling for permutation symmetry across seeds, the per-window weight vectors trace a coherent trajectory through a low-dimensional subspace, and an HMM fit to that trajectory recovers a set of latent states that partition the timeline into coherent phases ($K=11$ for Fakeddit, $K=16$ for Yelp, 15 visited). Second, models generalize better within states than across boundaries, an advantage that persists after controlling for temporal proximity. Third, the recovered states track the data's class distribution, operationalized as Jensen-Shannon divergence, more tightly than they track the weight-space distances that defined them. These results support the claim that the temporal structure of a drifting data stream is recoverable from the weights of the models it gives rise to, and membership in that structure predicts generalization in a manner that parallels the within-event coherence of human memory.

That every effect is attenuated on Yelp is informative about the differences between the datasets. On Fakeddit, the class distribution is highly unstable (Section~\ref{sec:temporal-windowing}), so there is large label-marginal drift for the states to latch onto. Yelp's star-rating distribution is more stable over time, leaving less label-marginal movement to structure the segmentation, which would account for the weaker JSD correlation and smaller within-state generalization advantage. Moreover, the drift that \citet{lukes-sogaard-2018-sentiment} document in review sentiment arises from lexical features shifting in polarity over time---a form of covariate shift that leaves the class distribution largely unchanged and would therefore be invisible to a JSD-based measure. 

While the difference between the HMM segmentation and the equal-size baseline was statistically significant at $p<0.0001$, the equal-size baseline was able to reproduce much of the within-state advantage on both datasets, with a difference of $D=+0.0051$ and $D=+0.0016$ on Fakeddit and Yelp respectively. We read this as a description of the data: each timeline is dominated by a few large regime shifts that any contiguous partition is likely to catch, hence the HMM's contribution is to refine where those few boundaries fall.
\subsection{Latent State Interpretation}
\begin{figure}[ht]\centering\includegraphics[width=0.75\linewidth]{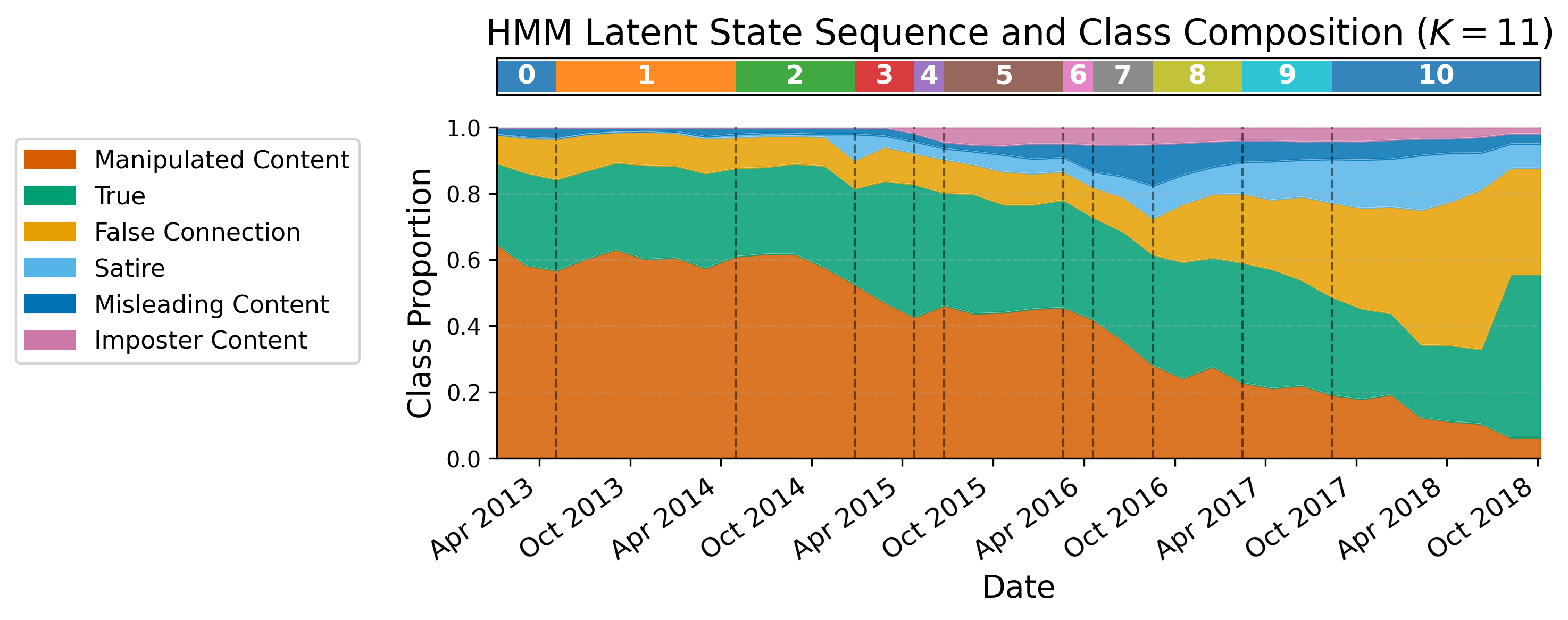}\caption{The 6-way class distribution across the 35 temporal windows in the Fakeddit dataset
(January 2013--October 2018), with the HMM-decoded state boundaries
($K=11$) overlaid.}\label{fig:dist_states}\end{figure}

The states were estimated from the model weights, yet the partition they induce aligns more closely with the underlying class distribution than with the Euclidean distance in the PCA space in which the HMM was fit. This ordering holds on both datasets, though the class distribution signal is far stronger on Fakeddit. Figure~\ref{fig:dist_states} overlays the recovered states on the Fakeddit class distribution across the 35 windows (Yelp overlay
given in Appendix~\ref{app:supp-yelp-dist}); the states appear to span stretches over which the class distribution holds comparatively steady, with boundaries falling at points of rapid shift. This is quantified in Section~\ref{sec:results-jsd-stability}: on both datasets, same-state window pairs are closer and less dispersed in class distribution than pairs spanning a boundary.

If the states were simply a regrouping of windows by class distribution, holding the distribution fixed would erase the within-state transfer advantage. However, the states predict transfer beyond class divergence and lag on both datasets, an effect that is substantial on Fakeddit and weak but statistically significant on Yelp (Section~\ref{sec:results-partial}), which indicates that the states also encode structure relevant to transfer that the label marginal does not express. The dominance of label shift as a signal is thus largely specific to Fakeddit. In the Yelp dataset, where the class distribution is more stable, the segmentation does not depend on strong label drift to recover a within-state advantage. 
\subsection{Limitations}

Our design fits a model to each non-overlapping window so that the weights are a function of a single period's data. However, real systems rarely operate this way. A deployed model typically accumulates data over time, and its state reflects its entire history rather than the most recent interval. This design was adopted deliberately, as a cumulative scheme would cause the training set to grow in size from window to window and introduce a confound between the quantity of data and its content. The cost of this control is that our regimes describe the data of each period instead of the trajectory of an accumulating learner. Additionally, human segmentation draws on world knowledge that a fixed-encoder classifier lacks. Misinformation detection, for instance, presupposes knowledge of what is true, whereas our models only see frozen embeddings. Knowledge-rich backbones such as large vision-language models \citep{Liu2024FKAOwl} or LLM-based sentiment classifiers \citep{Krugmann2024} could supply this world knowledge and are a natural extension.

The retained components capture only a small fraction of the total weight
variance---$11.47\%$ for Fakeddit, $11.10\%$
for Yelp. Although the post-alignment projection demonstrates that this is
largely sufficient to carry the temporal signal, most of the variation is left
unmodeled. Additionally, the Hungarian alignment removes the permutation of the
hidden layer, but not other symmetries of the network's reparameterization, such
as the positive-rescaling invariance of ReLU units. This could inflate the
distances we compute in weight space and offer a plausible explanation for why
the weight-side correlation underperforms the data-side correlation. The left-to-right topology forbids a latent
regime, once left, from recurring. This encodes the event-segmentation
assumption of \citet{Baldassano2017}, but it is a strong constraint. Seasonal effects in review sentiment or recurring content types on Reddit could cause the drifting stream to return to earlier regimes, which our topology would be forced to model as a new state. Forbidding recurrence also forces the segmentation to be
contiguous, so the equal-size baseline differs from the HMM only
in where boundaries fall, not in whether the partition is contiguous, limiting the comparison.

Because label-distribution shift is directly measurable and a strong predictor
of transfer, a natural objection is that one could segment the
timeline from the class distributions alone and dispense with the weight
extraction and HMM machinery. The partialling analysis in
Section~\ref{sec:results-partial} is evidence that the weight-derived states are
not reducible to a class distribution segmentation on either dataset. This applies especially to the Yelp dataset, where the JSD
correlation is weaker. However, fully establishing the method's
distinctive value requires additional evaluation on datasets in which the relevant drift is not known in
advance, or in which covariate shift invisible to the class distribution drives
the change. In these settings, event segmentation acts as an annotation-free, model-derived probe that offers something beyond direct data statistics.
\subsection{Future Work}
The correlation analyses established that the recovered states track the class
distribution more closely than the geometry of the weights. A natural next step
is to separate the drift that boundaries respond to into label-marginal and feature covariate shift, and to relate states to other data characteristics, such as
the volume, topical composition, or lexical content of each window. Additionally, our account
is descriptive, locating boundaries without explaining what drives them. A
mechanistic follow-up would tie transitions to changes
in the world the stream reflects, such as new content types or external events.
A segmentation that marks discontinuities in generalization is also a candidate
signal for when a deployed model should be retrained, connecting our descriptive
account to the predictive boundary-tracking of \citet{Won2018}.

The disjoint-window scheme could be replaced by a cumulative one, in which each
model is trained on all data up to a given time, testing whether boundaries
survive the integration of history. Allowing the encoders to train rather than
freezing them would allow the representation itself to drift and test whether
the segmentation sharpens. Our use of a hidden Markov model was motivated
primarily by its close connection to the event segmentation literature
\citep{Baldassano2017}, but future work could explore alternative approaches,
such as constrained k-means clustering and change-point detection, which may
recover different notions of event structure and make different assumptions
about boundary formation. The dimensionality reduction also need not be linear; manifold-learning methods such as T-PHATE \citep{Busch2023} could better capture nonlinear structure in the weight trajectory.

Our framework is not tied to weights. Any sequence of model-internal
representations elicited by an ordered input can serve as the HMM's emissions,
and richer choices may bring the setup closer to its neuroscientific source. Fitting the HMM to a model's activations as it processes the
same naturalistic stimuli used in human studies, one could compare the resulting
boundaries to the event boundaries reported by \citet{Baldassano2017}. This
would test whether the recovered event structure reflects input properties that
any capable learner must discover, or whether it is specific to human
perception.
\section{Conclusion}
\label{sec:conclusion}
Our study investigates whether the event-segmentation account of human
perception extends to an artificial learner, substituting a classifier's weights for cortical activity. On both datasets, an HMM fit to the weight trajectory recovered coherent regimes ($K=11$
for Fakeddit, $K=16$ for Yelp, 15 visited), models generalized better within states than across boundaries, and the
advantage survived controls for temporal proximity and a naive equal-size
partition. The recovered states align more closely with shifts in the data's class distribution than with distances in the weight space that defined them, yet the within-state signal survives after class divergence and temporal proximity
are partialled out---substantially on Fakeddit, and slightly but significantly on Yelp. We leave the method's extension to more general settings to future work.
\bibliographystyle{tmlr}
\bibliography{references}
\appendix
\section{Replication of Stepanova and Ross (2023)}
\label{app:replication}

This appendix collects the architecture, training protocol, and full results of
our replication of
\citet{stepanova-ross-2023-temporal}. Their study examined how the performance of
multimodal misinformation classifiers degrades when evaluated on temporally
out-of-domain data.

\subsection{Classifier Architecture}
\label{app:replication-arch}
The replication classifier follows the multimodal architecture of
\citet{stepanova-ross-2023-temporal}. Text titles are encoded with a
RoBERTa-family model, \texttt{all-distilroberta-v1} \citep{Sanh2019DistilBERT},
which outputs 768-dimensional embeddings. Images are encoded with the
2048-dimensional penultimate layer of a ResNet-50 backbone \citep{he2016deep}.
These two embeddings are concatenated into a 2816-dimensional vector and passed
to a one-hidden-layer feed-forward network with ReLU activation and a 6-way
output layer. We freeze the modality encoders and precompute the embeddings so
that only the MLP head is trained.

\subsection{Training Protocol and Hyperparameter Search}
\label{app:replication-training}
The training protocol used for the replication
of \citet{stepanova-ross-2023-temporal} differs
from the regime used to train the per-window models in the main experiments. In
particular, our replication follows the original study in using early stopping. The replication selects the hidden width $n$ and learning rate by a grid search over $n = 2^i$ for $i \in \{5, \dots, 14\}$ and
$\text{lr} \in \{10^{-2}, \dots, 10^{-5}\}$, with the setting maximizing
validation accuracy retained. Models are trained for a maximum of 20 epochs with
early stopping if validation accuracy does not improve over 4 consecutive
epochs.

Models are trained under two regimes whose comparison isolates the performance
gap attributable to temporal distribution shift. In the \quotes{original} (OG)
split, train, validation, and test sets are drawn at random from the full time
range. In the \quotes{temporal} split, the train, validation, and test sets
cover non-overlapping, chronologically ordered time periods, so that the model
is evaluated on a chronologically later test split. Following the data
preprocessing conventions of \citet{stepanova-ross-2023-temporal}, only posts
that contain an image are retained, so every sample used is multimodal.

\subsection{Results}
\label{app:replication-results}

\begin{table}[ht]
\centering
\caption{Experiment replication results on the held-out test set.}
\begin{tabular}{|l|c|c|}
\hline
\textbf{Condition} & \textbf{Micro F1} & \textbf{Macro F1} \\
\hline
OG 2-way     & 0.886 & 0.881 \\
Temporal 2-way & 0.853 & 0.820 \\
\hline
OG 6-way     & 0.849 & 0.737 \\
Temporal 6-way & 0.830 & 0.673 \\
\hline
\end{tabular}
\label{tab:exp1}
\end{table}

The results of the replication are reported in Table~\ref{tab:exp1}. Moving from the original to the temporal split produced consistent degradation
across both granularities. For 2-way classification, micro F1 fell by 3.3
percentage points (0.886 to 0.853) and macro F1 by 6.1 points (0.881 to 0.820).
For the 6-way task, micro F1 dropped by 1.9 points (0.849 to 0.830) and macro F1
by 6.4 points (0.737 to 0.673). The larger decrease in macro F1 than micro F1 in
both cases reflects disproportionate degradation on the minority classes,
consistent with the findings of \citet{stepanova-ross-2023-temporal}. Our
replicated values exceed those reported in the original study (2-way temporal:
micro 0.81, macro 0.78; 6-way temporal: micro 0.72, macro 0.52), although the
direction and magnitude of the temporal performance gap are consistent with the
original findings.
\section{Methodological Formalism}
\label{app:methods-formalism}

This appendix collects the formal definitions deferred from
Section~\ref{sec:methods}. Each subsection expands a step summarized in the
main text.

\subsection{Permutation Alignment}
\label{app:formalism-align}
This expands the alignment procedure of Section~\ref{sec:weight_processing}.
Hidden units in a feed-forward network are symmetric under permutation
\citep{Entezari2022}, so weight vectors must be aligned across seeds before
they can be compared. Each hidden unit $k$ is assigned an alignment signature
formed by concatenating its incoming weights (row of $W_h$) with its outgoing
weights (column of $W_o$):
\begin{equation}
g_k = \big[(W_h)_{k,:} \,\Vert\, (W_o)_{:,k}^{\top}\big] \in \mathbb{R}^{d+c},
\end{equation}
where $d$ is the input feature dimension and $c$ the number of classes, so that
the signature has length $d + c$ ($2822$ for Fakeddit, $773$ for Yelp). For a
given window, all seeds' models are aligned to a synthetic centroid by
repeated linear-assignment (Hungarian) matching \citep{Ainsworth2023}. A
centroid $\{\bar g_k\}$ is computed from the current set of all seeds' weight
matrices, and each seed $s$ is then re-matched to that centroid by minimizing
the total squared signature distance,
\begin{equation}
\pi^{(s)} = \arg\min_{\pi \in \mathcal{S}_{n}}
\sum_{k=1}^{n} \big\| g_k^{(s)} - \bar g_{\pi(k)} \big\|_2^2,
\end{equation}
with $\mathcal{S}_{n}$ the set of permutations on the $n$ hidden units ($n=1024$
for Fakeddit, $n=256$ for Yelp). The centroid is recomputed and the procedure
iterates to convergence. Each recovered permutation $\pi^{(s)}$ is applied
consistently across $W_h$, $b_h$, and $W_o$; the output biases $b_o$, which
carry no hidden-unit indexing, are left unchanged.
\subsection{The Distance-Conditioned Gap Statistic}
\label{app:formalism-statistic}
This expands the test statistic of Section~\ref{sec:evaluation}. Let $\mathcal{W}_d$ and $\mathcal{A}_d$ denote the within- and across-state pairs at
lag $d = |i-j|$, with counts $n^{w}_d = |\mathcal{W}_d|$ and
$n^{a}_d = |\mathcal{A}_d|$, and let $\bar F^{w}_d$, $\bar F^{a}_d$ be the
corresponding mean macro F1 scores. The per-lag gap is
$\Delta_d = \bar F^{w}_d - \bar F^{a}_d$, and the test statistic is the
harmonic-count-weighted average
\begin{equation}
T = \frac{\sum_d w_d\, \Delta_d}{\sum_d w_d},
\qquad
w_d = \frac{2\, n^{w}_d\, n^{a}_d}{\,n^{w}_d + n^{a}_d\,}.
\end{equation} The harmonic mean of the two counts is near zero whenever either count is small, hence a lag is down-weighted unless it is well populated in both groups, in which case $\Delta_d$ is reliably estimated. 
\subsection{Pooled Gap Statistics for the Equal-Size Control}
\label{app:formalism-equalsize}
This expands the aggregation and difference statistics of
Section~\ref{sec:equal-size}. For the difference test, the pooled gap of a
segmentation is
\begin{equation}
\Pi = \bar F^{w} - \bar F^{a},
\end{equation}
the mean within-group F1 minus the mean across-group F1 over all off-diagonal
pairs, and the test statistic is the difference of pooled gaps between the two
segmentations,
\begin{equation}
D = \Pi_{\mathrm{HMM}} - \Pi_{\mathrm{equal}}.
\end{equation}
The null distribution for $D$ is formed by freely permuting the per-window HMM
state labels while preserving the number of windows per state, recomputing
$\Pi_{\mathrm{HMM}}$, and holding $\Pi_{\mathrm{equal}}$ fixed. Note that this per-window permutation is different from the lag-stratified shuffle that was used to analyze the segmentations individually. A lag-stratified shuffle would only scramble the within/across labels of a fixed segmentation, reproducing the single-segmentation analysis. Permuting the per-window labels instead generates a null distribution of alternative segmentations, assessing whether the HMM boundaries produce a larger within-group advantage over the equal-size baseline than random regroupings of the same windows.
\subsection{Pairwise Class Distribution Stability}
\label{app:formalism-stability}
This expands the within- versus across-state stability analysis of
Section~\ref{sec:stability}. For each unordered pair of windows $(i, j)$, we
compute the Jensen-Shannon divergence between their class distributions,
$\mathrm{JSD}(c_i \,\Vert\, c_j)$, and assign the pair to the within-group if
windows $i$ and $j$ share an HMM state and to the across-group otherwise. Let
\begin{equation}
\mathcal{W} = \{(i,j) : i < j,\, s_i = s_j\},
\qquad
\mathcal{A} = \{(i,j) : i < j,\, s_i \neq s_j\}
\end{equation}
denote the within- and across-state pairs respectively, where $s_i$ is the
HMM-decoded state of window $i$. The group mean divergences are
\begin{equation}
\bar{D}^{w} = \frac{1}{|\mathcal{W}|} \sum_{(i,j)\in\mathcal{W}}
\mathrm{JSD}(c_i \,\Vert\, c_j),
\qquad
\bar{D}^{a} = \frac{1}{|\mathcal{A}|} \sum_{(i,j)\in\mathcal{A}}
\mathrm{JSD}(c_i \,\Vert\, c_j).
\end{equation}
We compare the two groups on the mean and variance of pairwise JSD and assess
the difference in mean pairwise JSD,
$\Delta_{\mathrm{JSD}} = \bar{D}^{a} - \bar{D}^{w}$, with a label-shuffle
permutation test. The within/across labels are permuted across all
$\binom{W}{2}$ window pairs over $10,000$ iterations, holding the JSD values
fixed, and the observed $\Delta_{\mathrm{JSD}}$ is compared to the resulting null
distribution to form a one-sided empirical $p$-value.
\subsection{Partial Permutation Tests for the Class Distribution Control}
\label{app:formalism-partial}
This expands the significance tests for the combined model of
Section~\ref{sec:partial-jsd}. The within-state coefficient
$\beta_{\mathrm{state}}$ is assessed with a Freedman-Lane partial permutation
test \citep{FreedmanLane1983}. We fit the reduced model
$\tilde{F}_{ij} \sim \mathrm{JSD}(c_i \,\Vert\, c_j) + d$, permute its residuals within
each lag stratum to preserve the temporal-proximity control, rebuild the
response from the reduced-model fit plus the permuted residuals, refit the full
model, and collect $\beta_{\mathrm{state}}$ over permutations to form the null
distribution.

As a complementary test, we residualize transfer F1 on divergence and lag,
taking $r_{ij} = \tilde{F}_{ij} - \hat{\tilde{F}}_{ij}$ where $\hat{\tilde{F}}_{ij}$ is the fitted value
of $\tilde{F}_{ij} \sim \mathrm{JSD}(c_i \,\Vert\, c_j) + d$, and apply the
harmonic-weighted, lag-stratified distance-conditioned gap statistic of
Section~\ref{sec:evaluation} to the residuals $r_{ij}$, comparing the
within-group and across-group residual means.
\section{Permutation Null Distributions}
\label{app:null_distributions}

All permutation null distributions referenced in the main text are collected here. In each figure, the grey histogram shows the distribution of the test statistic under the null hypothesis across 10,000 permutation iterations, and the red vertical line marks the observed statistic.

\subsection{Within- vs.\ Across-State Generalization}
\label{app:null_wa}

\begin{figure}[H]
\centering
\begin{subfigure}[t]{0.48\textwidth}
    \centering
    \includegraphics[width=\linewidth]{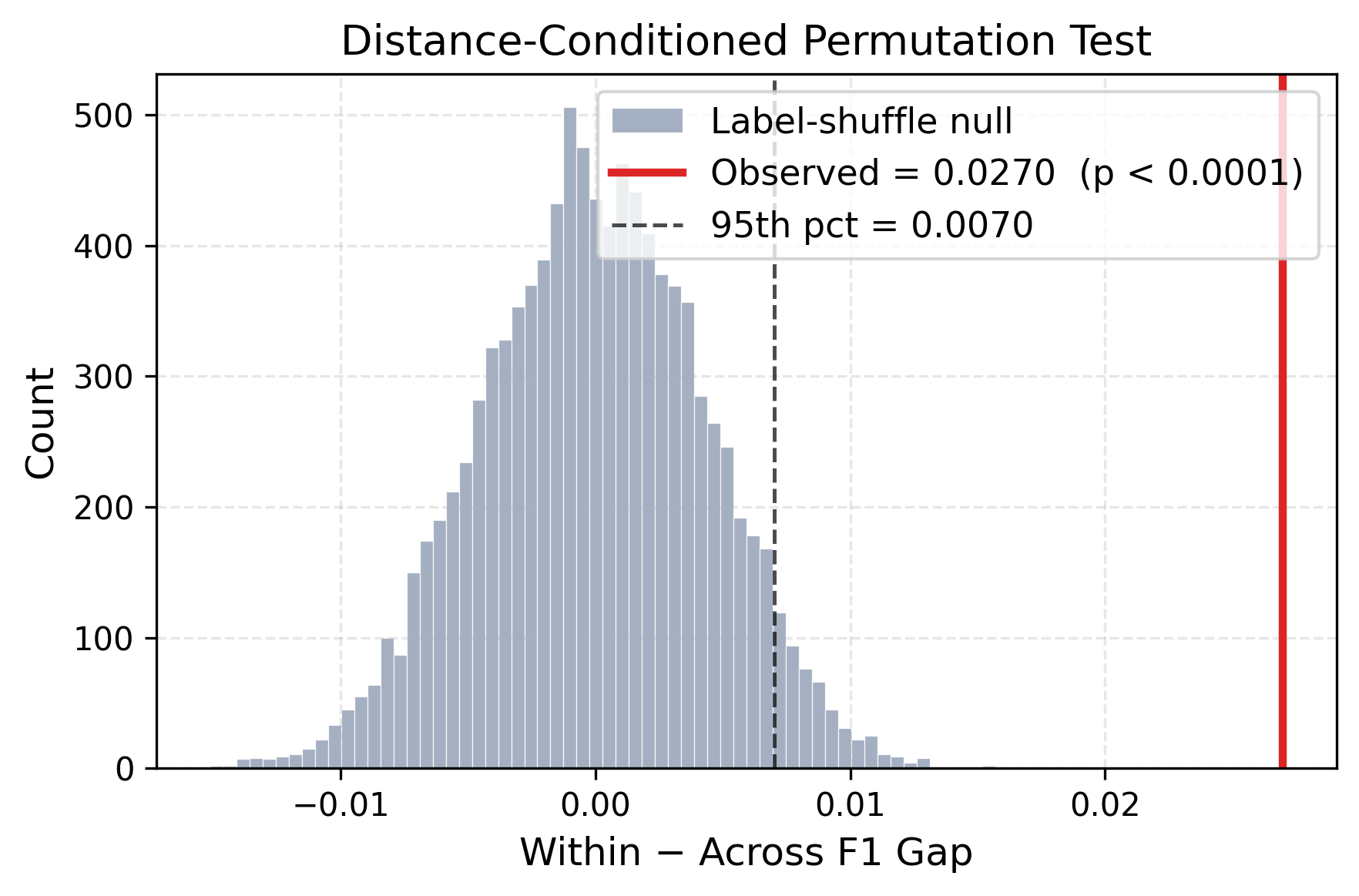}
    \caption{Fakeddit: observed statistic $+0.0270$ ($p < 0.0001$).}
    \label{fig:perm_null_fakeddit}
\end{subfigure}
\hfill
\begin{subfigure}[t]{0.48\textwidth}
    \centering
    \includegraphics[width=\linewidth]{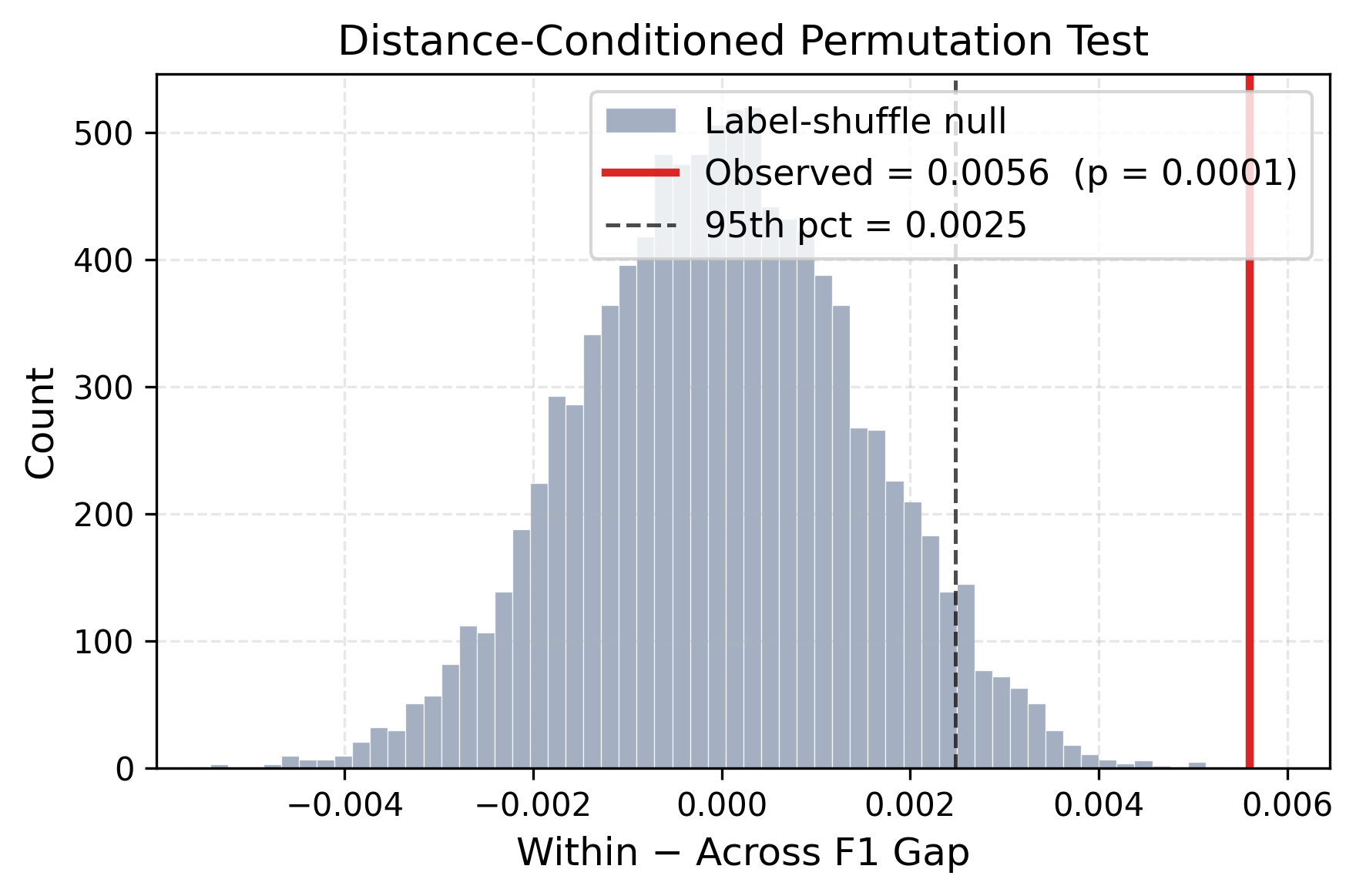}
    \caption{Yelp: observed statistic $+0.0056$ ($p = 0.0001$).}
    \label{fig:perm_null_yelp}
\end{subfigure}
\caption{Null distribution of the distance-conditioned gap statistic
under 10,000 label-shuffle permutations. The observed statistic (red
line) falls well outside the null distribution for both datasets.}
\label{fig:perm_null}
\end{figure}

\subsection{Comparison Against Equal-Size Segmentation}
\label{app:null_equal}
\begin{figure}[H]
\centering
\begin{subfigure}[t]{0.48\textwidth}
    \centering
    \includegraphics[width=\linewidth]{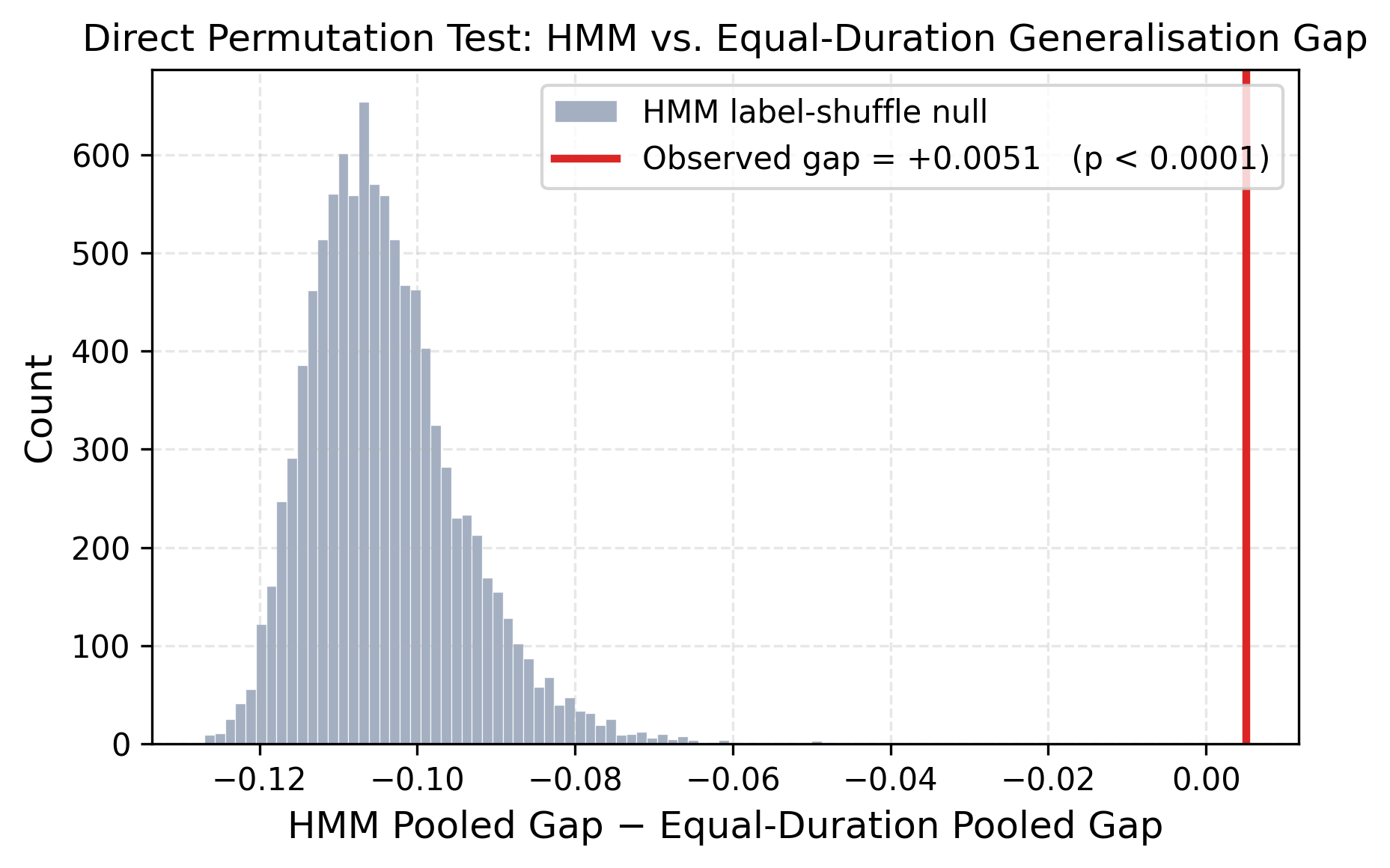}
    \caption{Fakeddit: observed difference $D = +0.0051$, $p < 0.0001$.}
    \label{fig:equal_perm_fakeddit}
\end{subfigure}
\hfill
\begin{subfigure}[t]{0.48\textwidth}
    \centering
    \includegraphics[width=\linewidth]{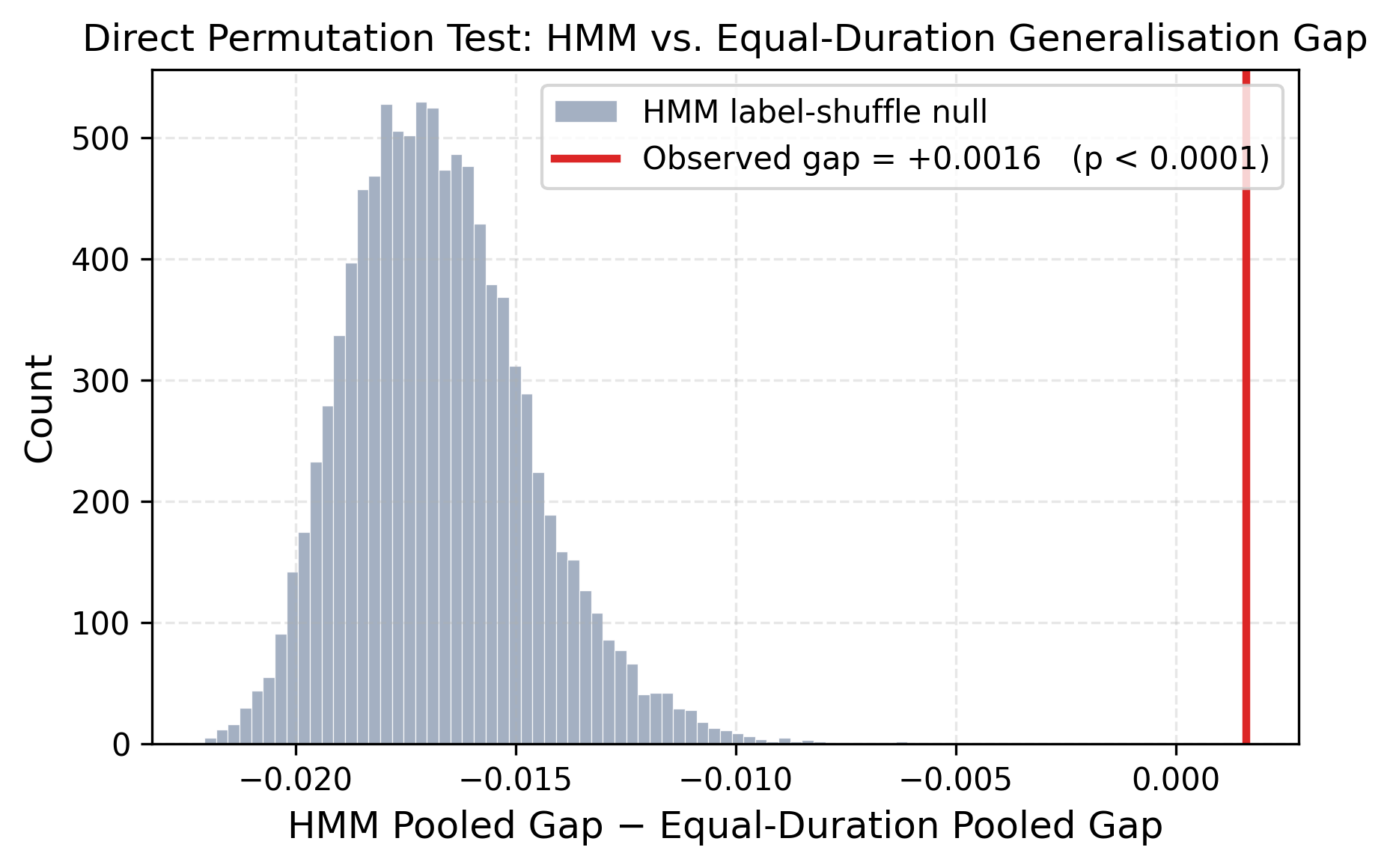}
    \caption{Yelp: observed difference $D = +0.0016$, $p < 0.0001$.}
    \label{fig:equal_perm_yelp}
\end{subfigure}
\caption{Null distribution of the HMM$-$equal-size gap difference
under 10,000 per-window label permutations. The observed difference
falls outside the null distribution for both datasets.}
\label{fig:equal_perm}
\end{figure}

\subsection{Correlation Analyses}
\label{app:null_corr}

\begin{figure}[H]
\centering
\includegraphics[width=\linewidth]{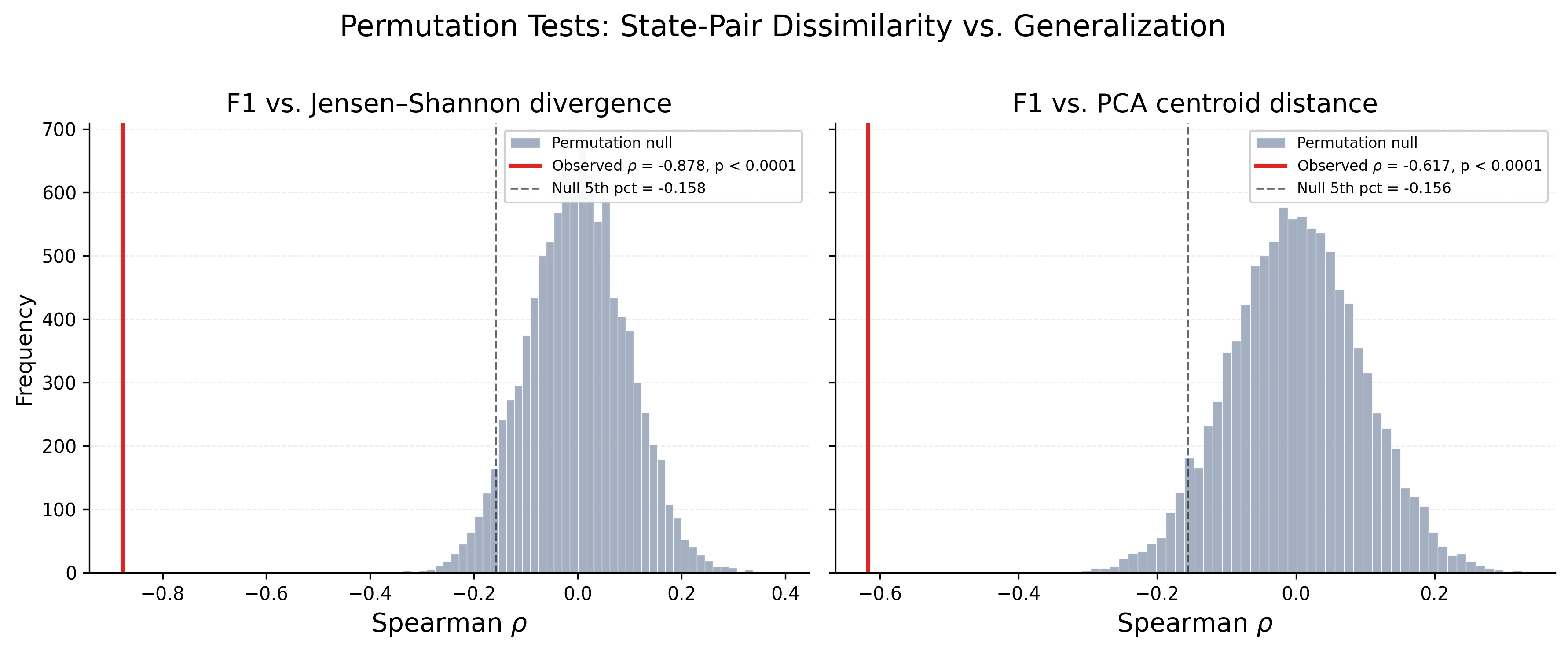}
\caption{Fakeddit: permutation null distributions (10,000 iterations) for the
two Spearman correlations between state-pair dissimilarity and mean
cross-window macro F1. \textit{Left}: class distribution (JSD).
\textit{Right}: weight-space centroid distance. Observed correlations
($\rho = -0.878$ and $\rho = -0.617$) both fall well outside their respective nulls
($p < 0.0001$).}
\label{fig:corr_analysis_fakeddit}
\end{figure}
\begin{figure}[H]
\centering
\includegraphics[width=\linewidth]{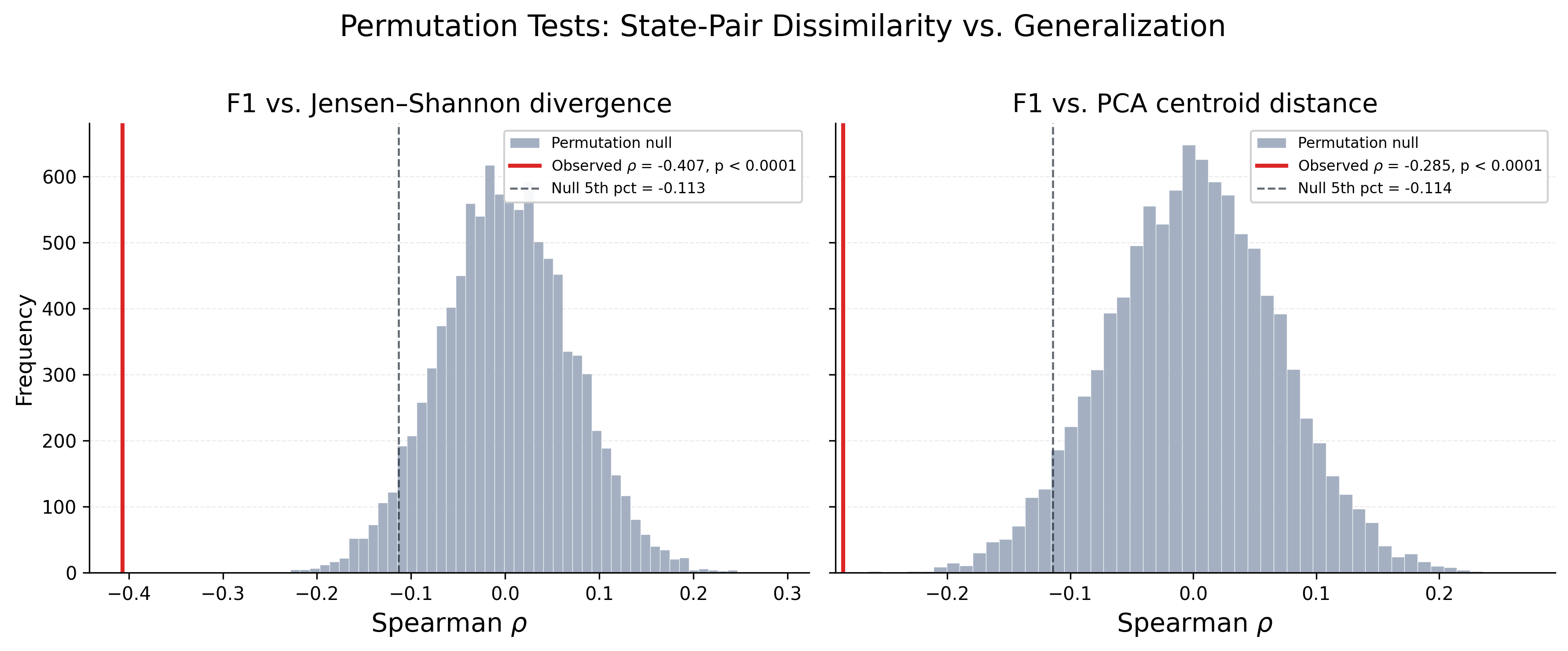}
\caption{Yelp: permutation null distributions (10,000 iterations) for the
two Spearman correlations between state-pair dissimilarity and mean
cross-window macro F1. \textit{Left}: class distribution (JSD).
\textit{Right}: weight-space centroid distance. Observed correlations
($\rho = -0.407$ and $\rho = -0.285$) both fall well outside their respective nulls
($p < 0.0001$).}
\label{fig:corr_analysis_yelp}
\end{figure}

\subsection{Within-State Class Distribution Stability}
\label{app:null_jsd}

\begin{figure}[H]
\centering
\begin{subfigure}[t]{0.48\textwidth}
    \centering
    \includegraphics[width=\linewidth]{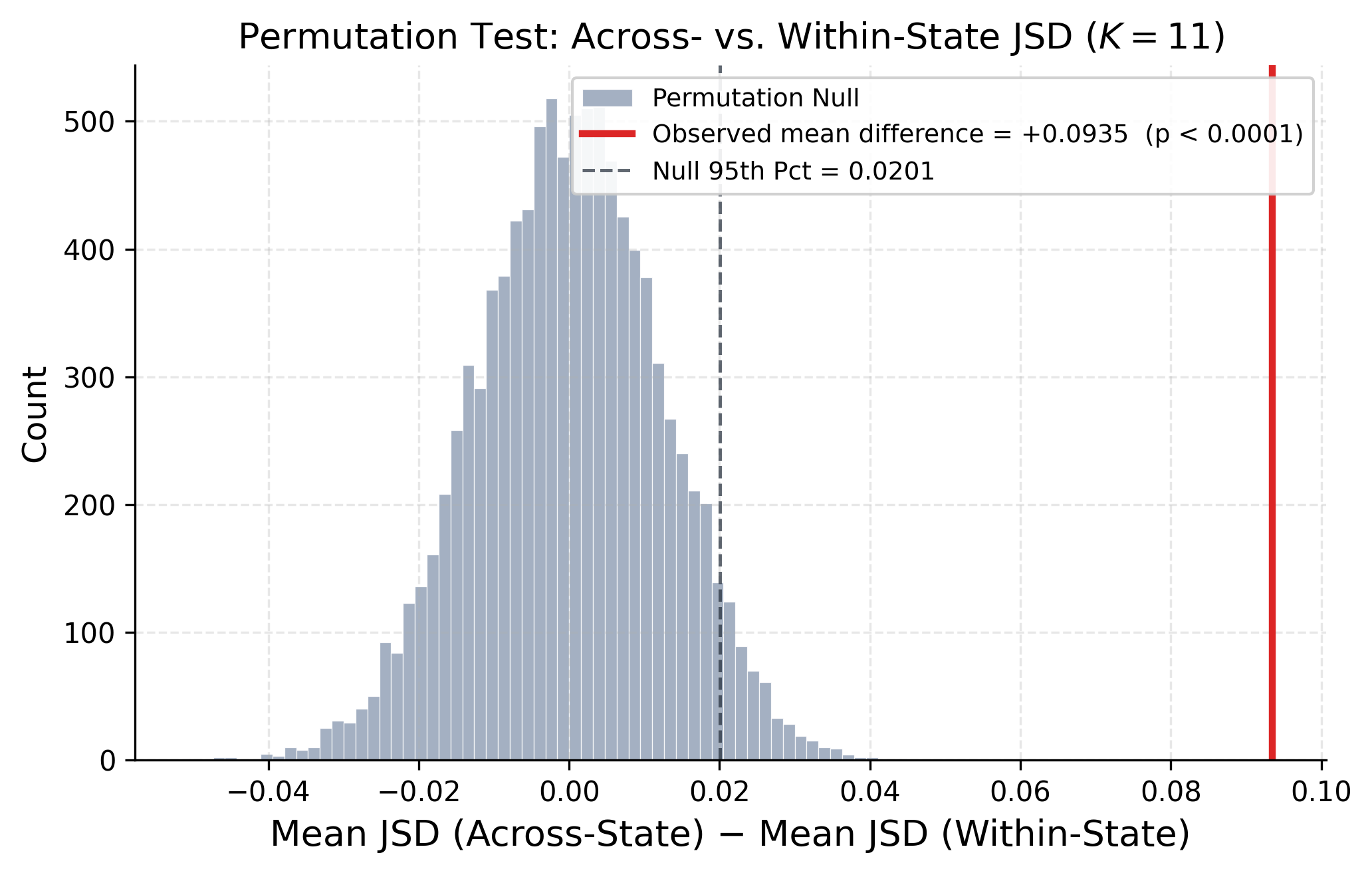}
    \caption{Fakeddit: observed difference $\delta = +0.0935$,
    $p < 0.0001$.}
    \label{fig:jsd_null_fakeddit}
\end{subfigure}
\hfill
\begin{subfigure}[t]{0.48\textwidth}
    \centering
    \includegraphics[width=\linewidth]{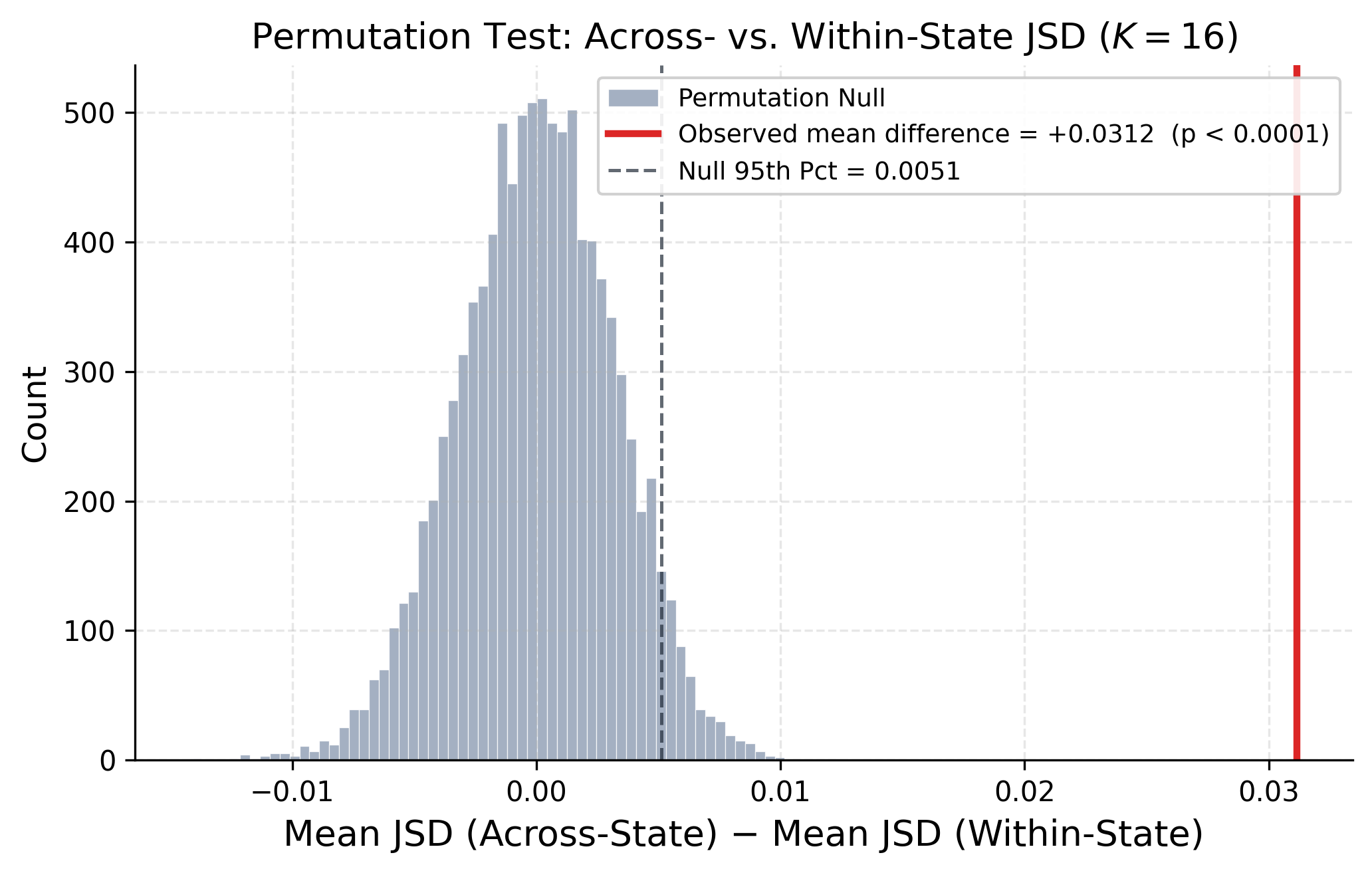}
    \caption{Yelp: observed difference $\delta = +0.0312$,
    $p < 0.0001$.}
    \label{fig:jsd_null_yelp}
\end{subfigure}
\caption{Null distribution of the difference in mean pairwise JSD
across- versus within-states under $10,000$ label-shuffle permutations. The
observed difference falls outside the null in both datasets.}
\label{fig:jsd_null}
\end{figure}

\subsection{State Membership Beyond Class Distribution}
\label{app:null_partial}

\begin{figure}[H]
\centering
\begin{subfigure}[t]{0.48\textwidth}
    \centering
    \includegraphics[width=\linewidth]{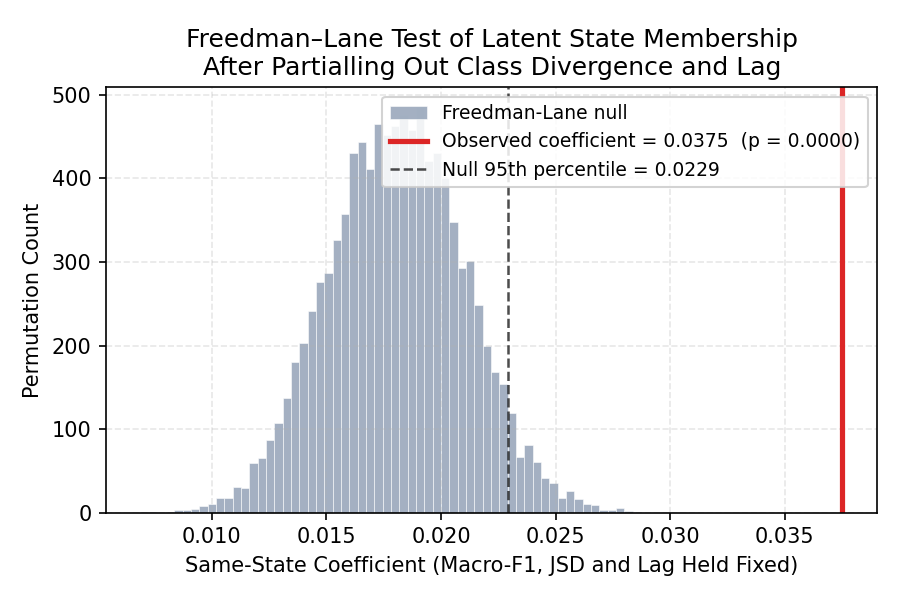}
    \caption{Fakeddit: observed coefficient $+0.0375$,
    $p < 0.0001$.}
    \label{fig:combined_null_fakeddit}
\end{subfigure}
\hfill
\begin{subfigure}[t]{0.48\textwidth}
    \centering
    \includegraphics[width=\linewidth]{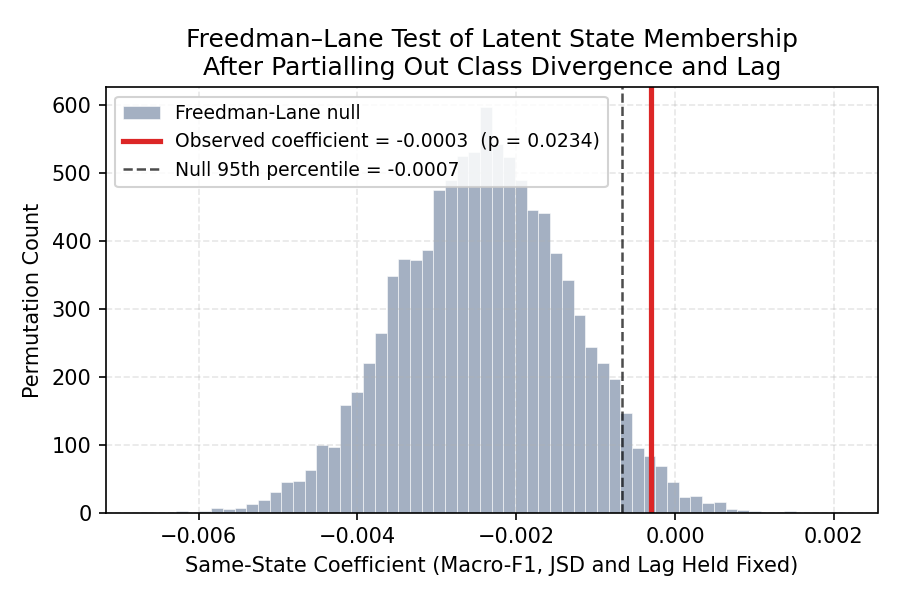}
    \caption{Yelp: observed coefficient $-0.0003$, $p = 0.0234$.}
    \label{fig:combined_null_yelp}
\end{subfigure}
\caption{Freedman-Lane null distribution for the same-state coefficient
$\beta_{\mathrm{state}}$ in the combined model, under $10,000$ lag-stratified
residual permutations. The observed coefficient falls in the upper tail of the
null on both datasets.}
\label{fig:combined_null}
\end{figure}
\begin{figure}[H]
\centering
\begin{subfigure}[t]{0.48\textwidth}
    \centering
    \includegraphics[width=\linewidth]{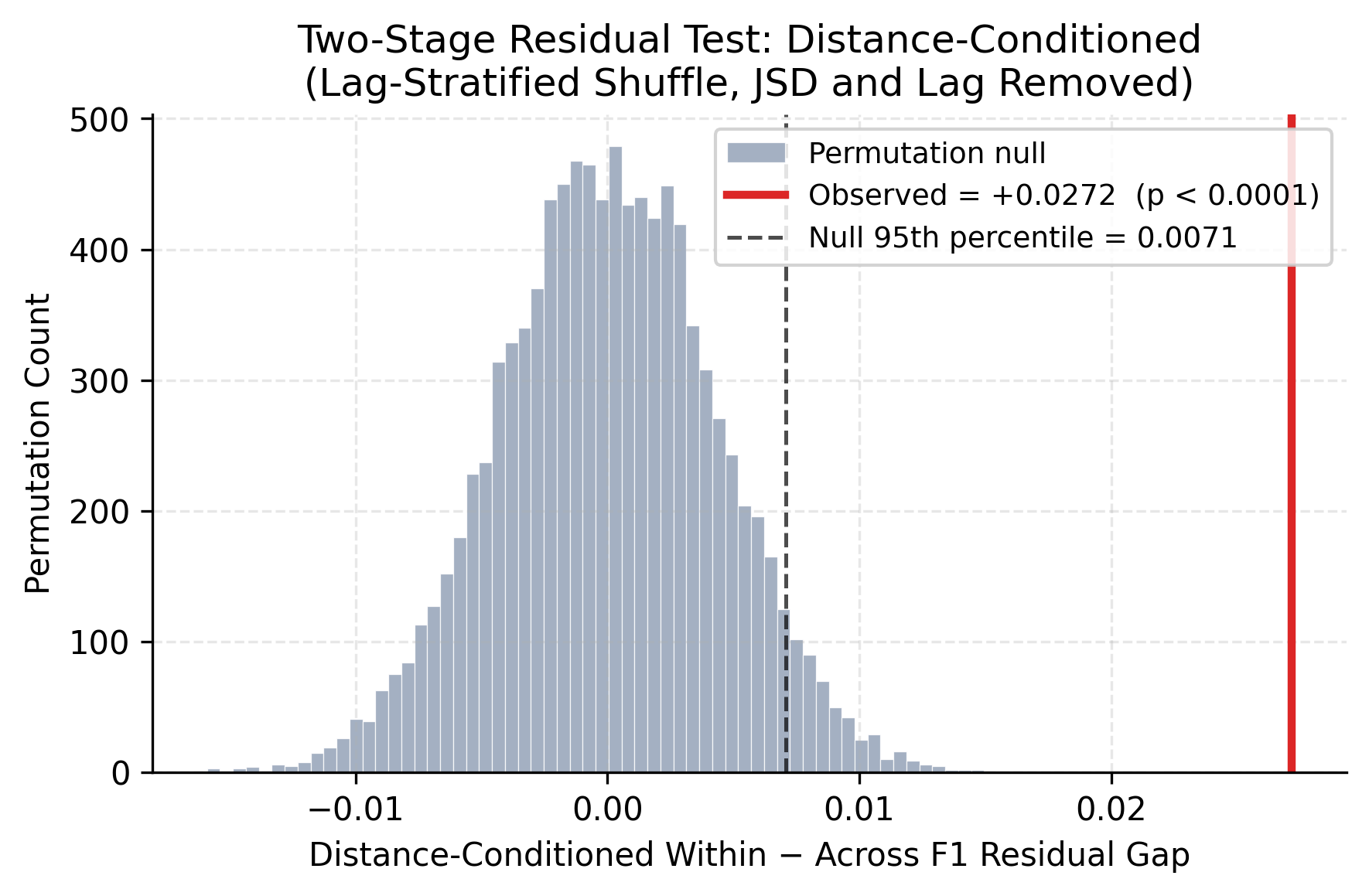}
    \caption{Fakeddit: observed gap $+0.0272$,
    $p < 0.0001$.}
    \label{fig:partial_resid_null_fakeddit}
\end{subfigure}
\hfill
\begin{subfigure}[t]{0.48\textwidth}
    \centering
    \includegraphics[width=\linewidth]{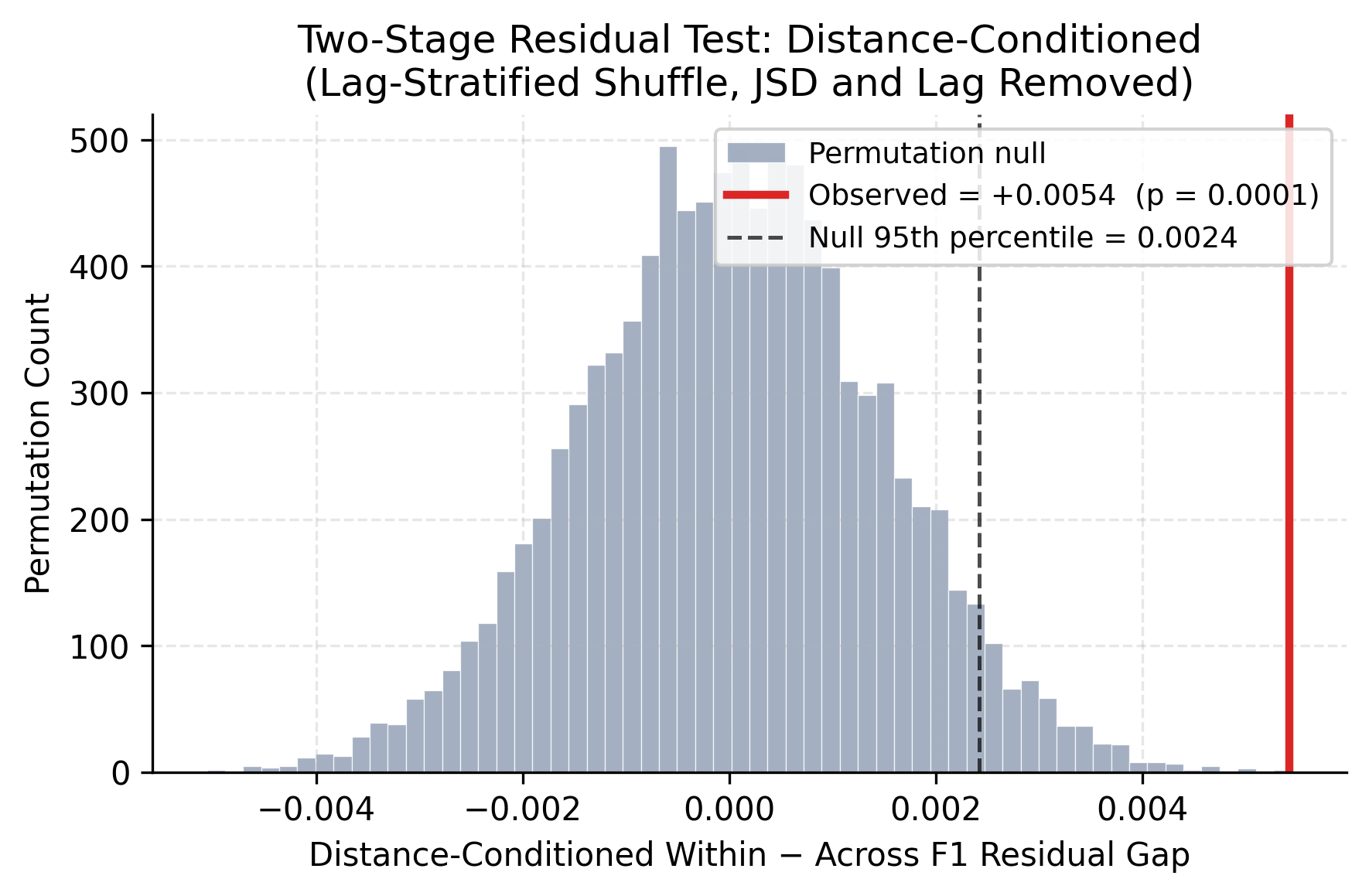}
    \caption{Yelp: observed gap $+0.0054$,
    $p = 0.0001$.}
    \label{fig:partial_resid_null_yelp}
\end{subfigure}
\caption{Null distribution of the lag-stratified, harmonic-weighted
distance-conditioned gap statistic applied to the divergence-and-lag-residualized
transfer F1, under $10,000$ permutations. The observed gap falls in the upper
tail of the null on both datasets.}
\label{fig:partial_resid_null}
\end{figure}
\section{Supplementary Figures}
\label{app:supp_figs}

This appendix collects figures that support choices and results stated
numerically in the main text. Each is referenced from the relevant section.
\subsection{Weight Space Before Permutation Alignment}
\label{app:supp-prealign}

\begin{figure}[H]
\centering
\begin{subfigure}[t]{0.49\textwidth}
    \centering
    \includegraphics[width=\linewidth]{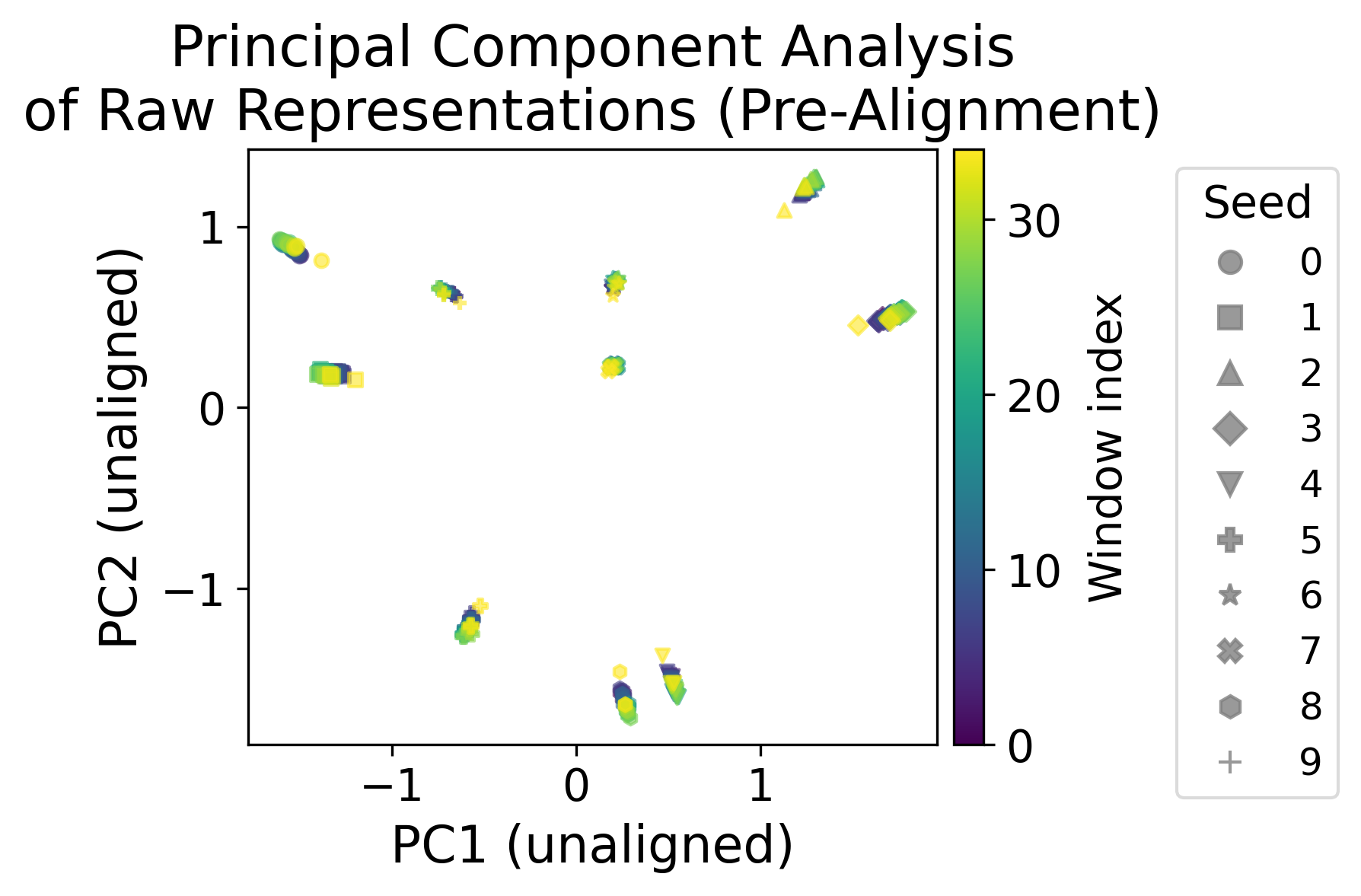}
    \caption{Fakeddit.}
    \label{fig:pca_space_before_fakeddit}
\end{subfigure}
\hfill
\begin{subfigure}[t]{0.49\textwidth}
    \centering
    \includegraphics[width=\linewidth]{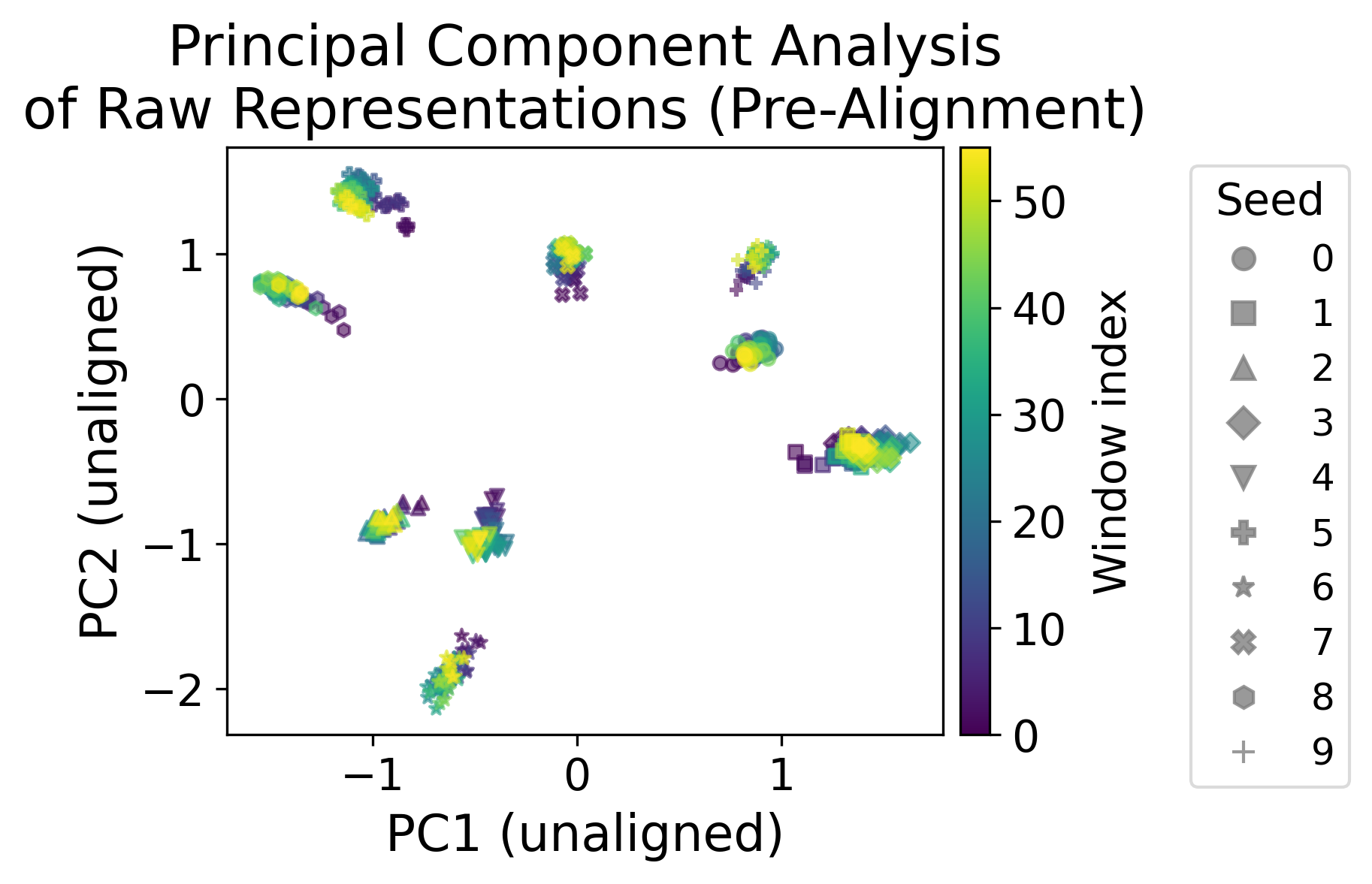}
    \caption{Yelp.}
    \label{fig:pca_space_before_yelp}
\end{subfigure}
\caption{PCA projections of the flattened MLP head weights (PC1 vs.\ PC2)
before permutation alignment, for (a) Fakeddit and (b) Yelp. Clustering is organized by seed rather than by
temporal window, reflecting the permutation symmetry of the hidden layer.}
\label{fig:pca_space_before}
\end{figure}
\subsection{Weight-Space Component Retention}
\label{app:supp-scree}

\begin{figure}[H]
\centering
\begin{subfigure}[t]{0.49\textwidth}
    \centering
    \includegraphics[width=\linewidth]{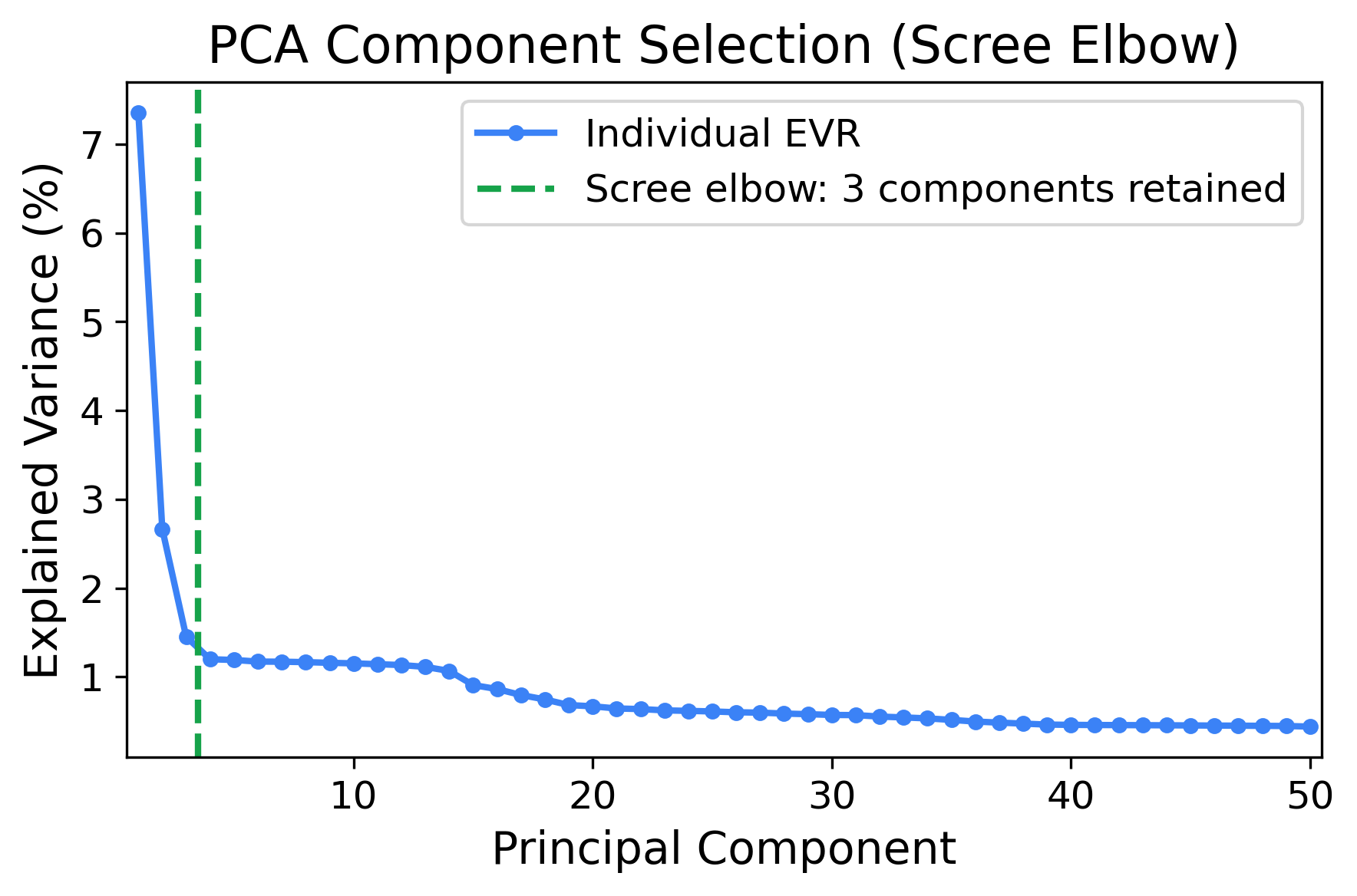}
    \caption{Fakeddit.}
    \label{fig:pca_scree_fakeddit}
\end{subfigure}
\hfill
\begin{subfigure}[t]{0.49\textwidth}
    \centering
    \includegraphics[width=\linewidth]{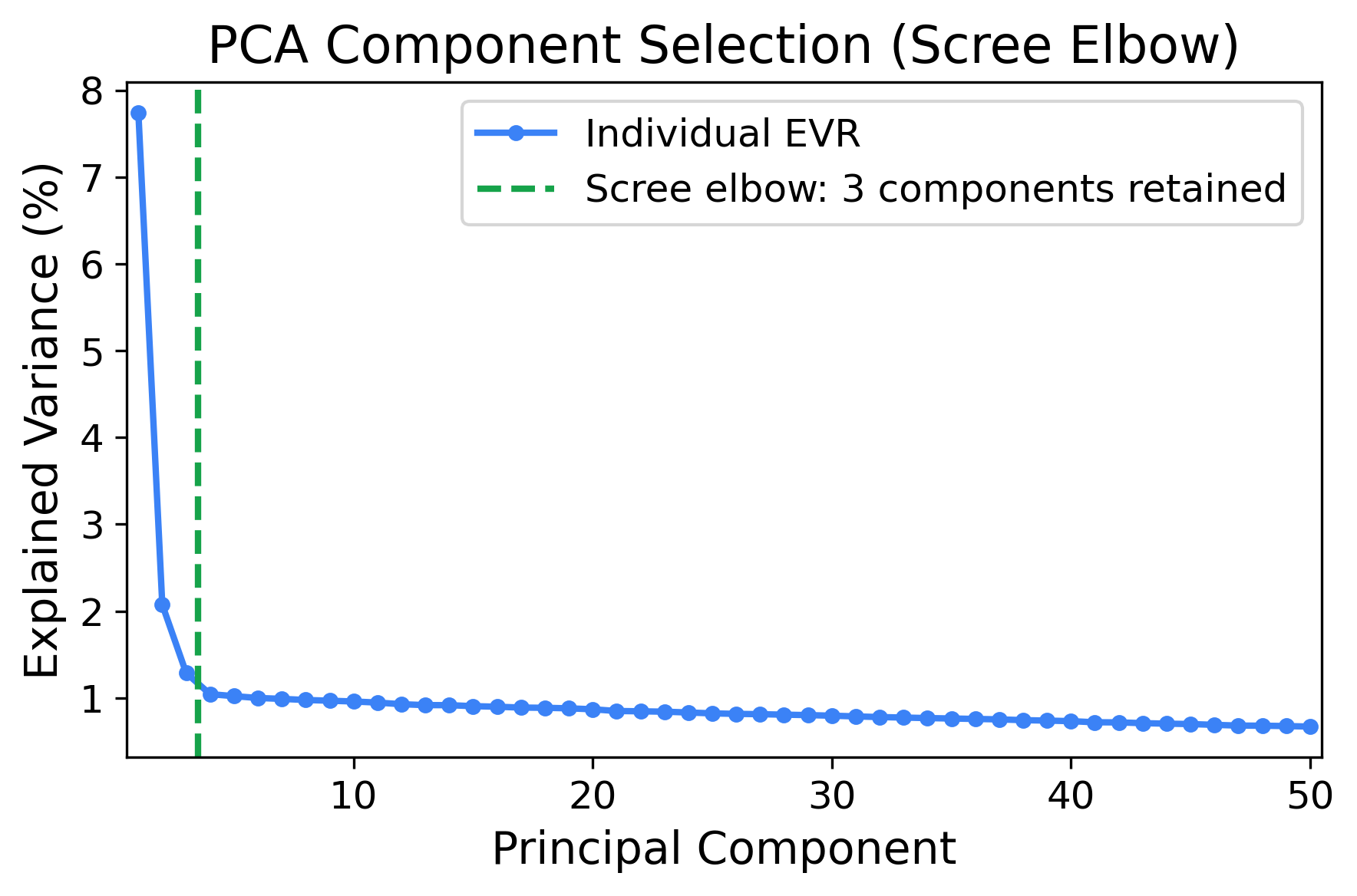}
    \caption{Yelp.}
    \label{fig:pca_scree_yelp}
\end{subfigure}
\caption{Per-component explained variance ratio for the first 50 principal
components of the aligned weight matrix, for (a) Fakeddit and (b) Yelp.
The vertical dashed line marks the retention cutoff at PC3, selected at
the elbow of the scree curve; the three components to its left are retained.}
\label{fig:pca_scree}
\end{figure}

\subsection{HMM Model Selection}
\label{app:supp-selection}

\begin{figure}[H]
\centering
\includegraphics[width=\linewidth]{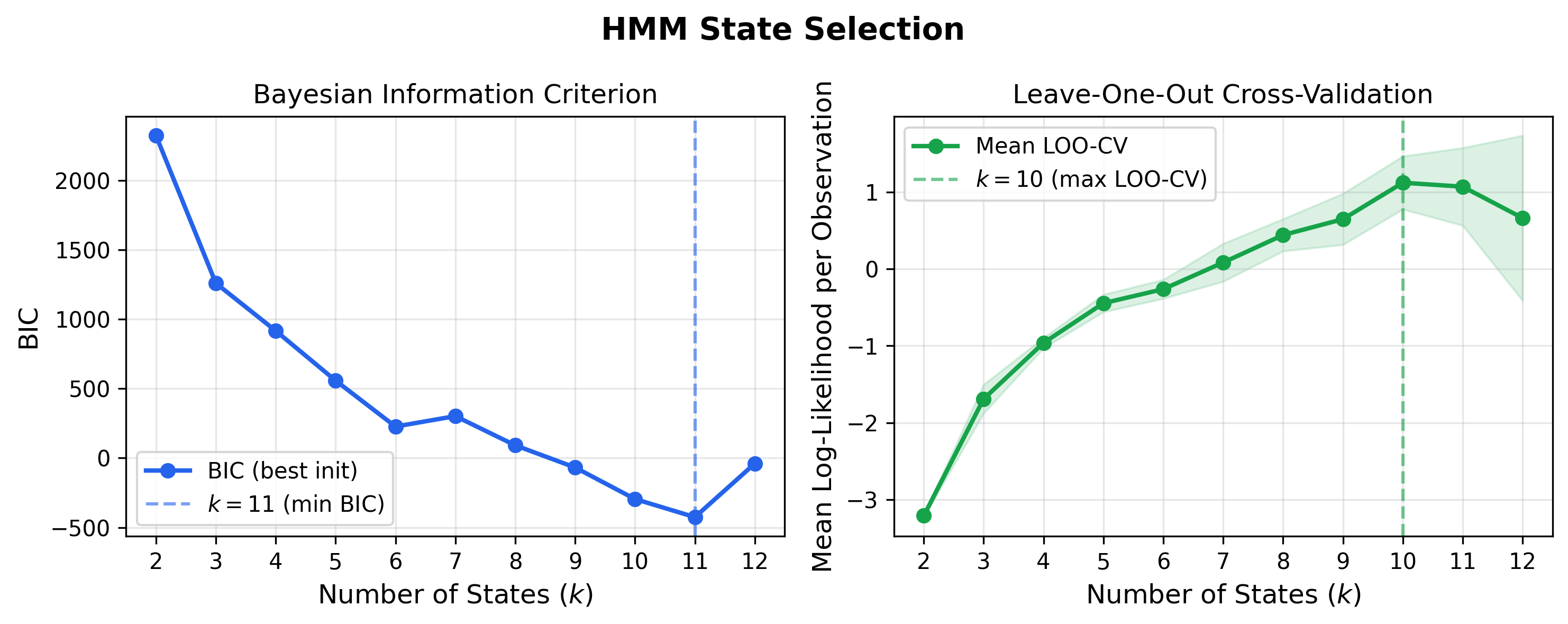}
\caption{HMM model selection for the Fakeddit dataset ($K=2$--$12$).
\textit{Left}: Bayesian information criterion, minimized at $K=11$
(BIC $\approx -425.67$). \textit{Right}: leave-one-seed-out
cross-validation log-likelihood per observation (mean $\pm$ 1 std
across folds). $K=11$ was selected.}
\label{fig:hmm_select_fakeddit}
\end{figure}

\begin{figure}[H]
\centering
\includegraphics[width=\linewidth]{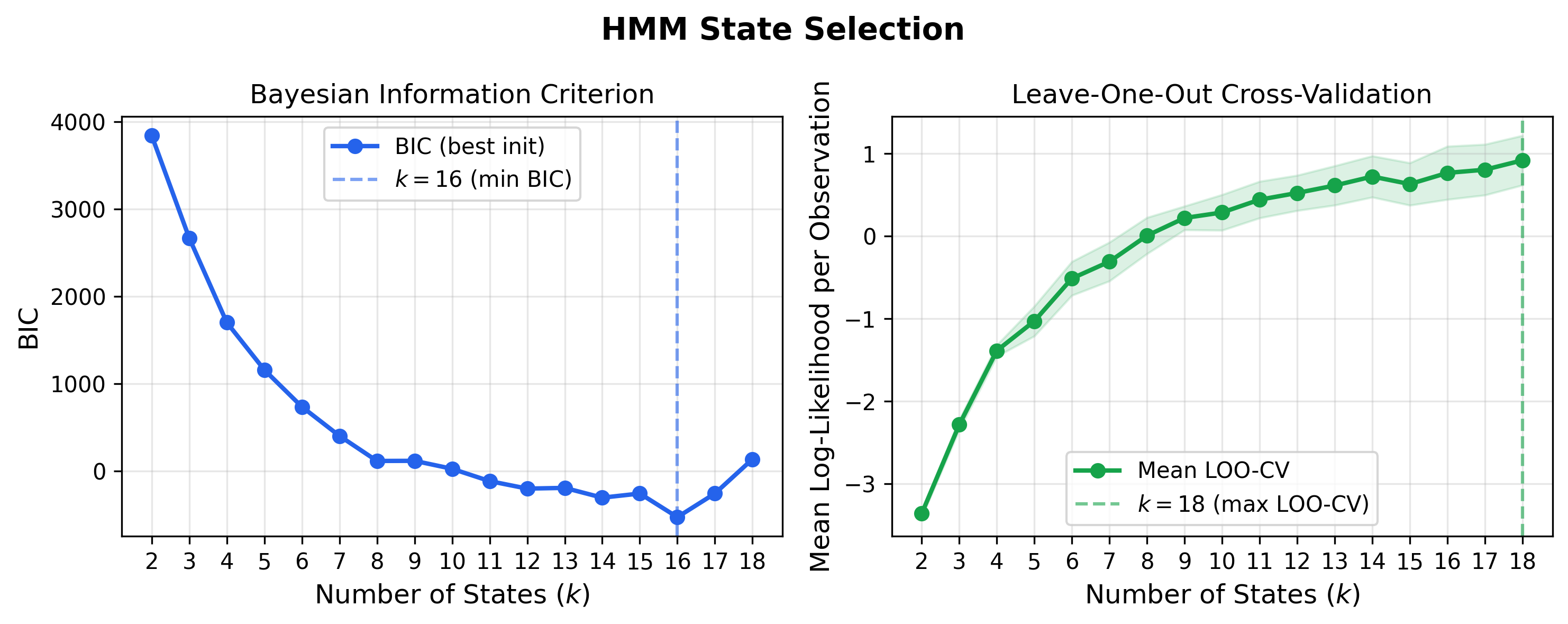}
\caption{HMM model selection for the Yelp dataset ($K=2$--$18$).
\textit{Left}: Bayesian information criterion, minimized at $K=16$
(BIC $\approx -527.11$). \textit{Right}: leave-one-seed-out
cross-validation log-likelihood per observation (mean $\pm$ 1 std
across folds). $K=16$ was selected.}
\label{fig:hmm_select_yelp}
\end{figure}

\subsection{Equal-Size Control}
\label{app:supp-eqlag}

\begin{figure}[H]
\centering
\includegraphics[width=\linewidth]{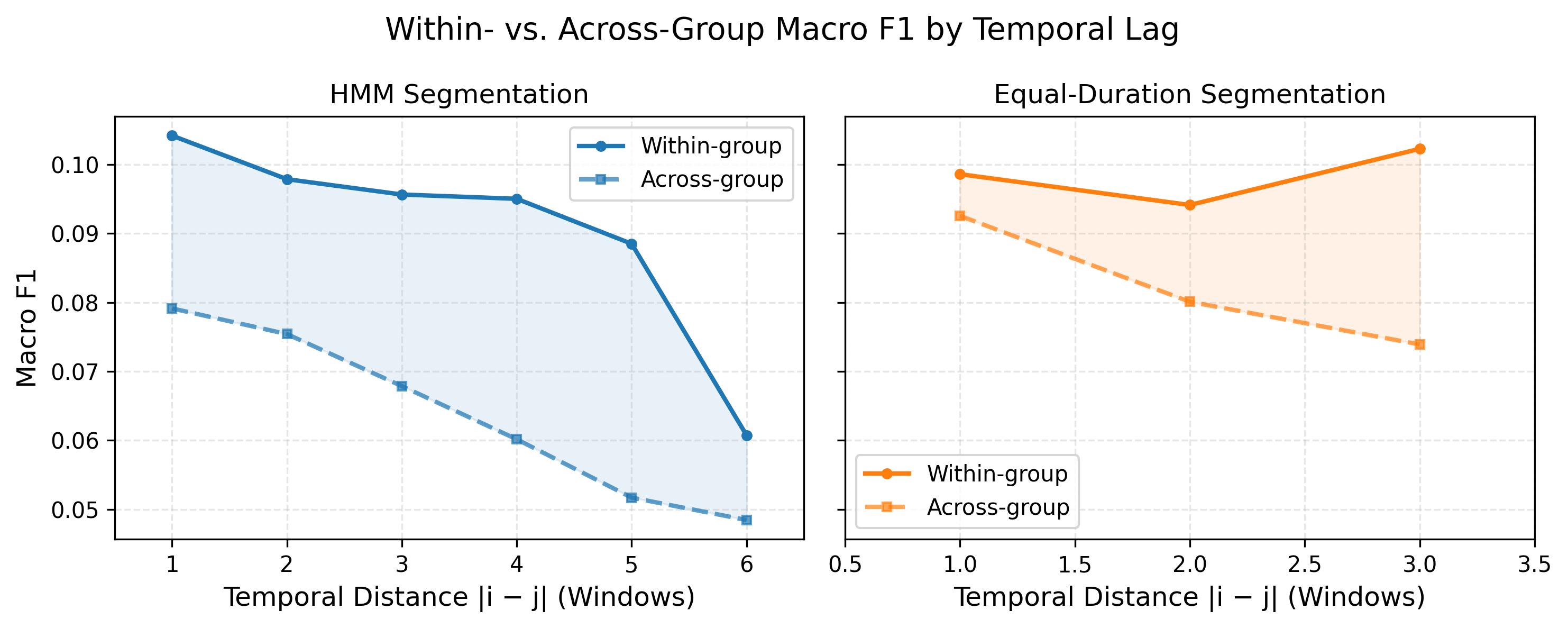}
\caption{Fakeddit: within- and across-group macro F1 as a function of temporal
lag for the HMM ($K=11$, left) and equal-size (right) segmentations. The
HMM within-group advantage is visible across lags, whereas the
advantage is weaker for the equal-size segmentation.}
\label{fig:f1_vs_distance_both_fakeddit}
\end{figure}

\begin{figure}[H]
\centering
\includegraphics[width=\linewidth]{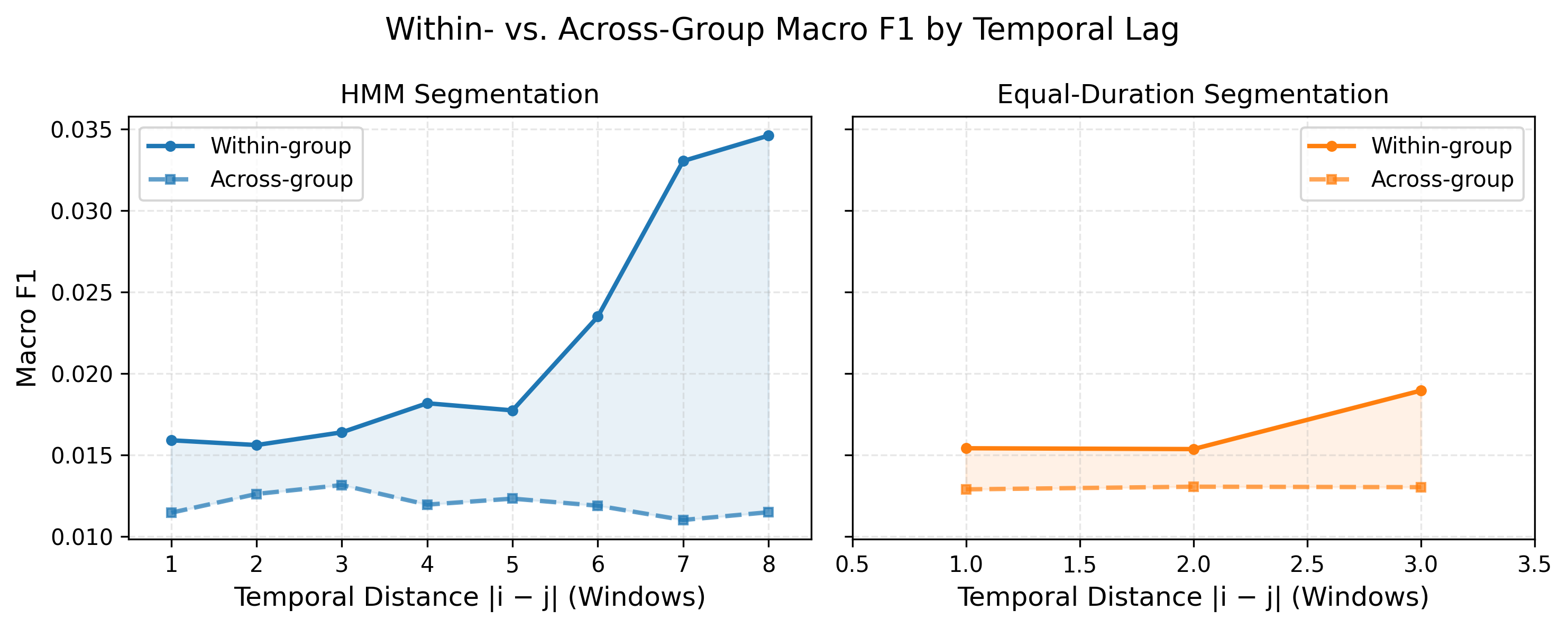}
\caption{Yelp: within- and across-group macro F1 as a function of temporal
lag for the HMM ($K=16$, left) and equal-size (right) segmentations. The
HMM within-group advantage is visible across lags, whereas the
advantage is weaker for the equal-size segmentation.}
\label{fig:f1_vs_distance_both_yelp}
\end{figure}
\subsection{Class Distribution and State Boundaries for Yelp}
\label{app:supp-yelp-dist}

\begin{figure}[H]
\centering
\includegraphics[width=0.75\linewidth]{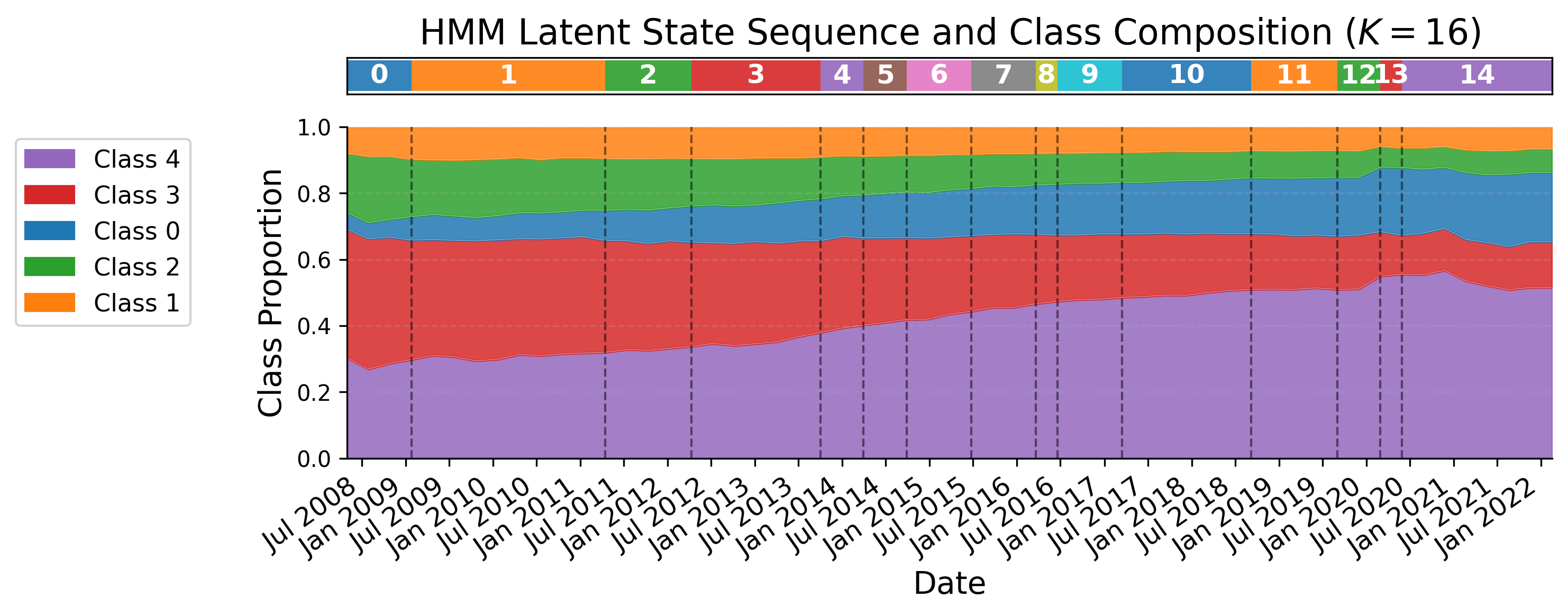}
\caption{The 5-way class distribution across the 56 temporal windows in the
Yelp dataset (May 2008--February 2022), with the HMM-decoded state
boundaries ($K=16$, 15 visited) overlaid. The star-rating distribution is
markedly more stable across windows than the Fakeddit class distribution
(Figure~\ref{fig:dist_states}), consistent with the attenuated
class-divergence effects reported in Sections~\ref{sec:results-corr}
and~\ref{sec:results-jsd-stability}.}
\label{fig:dist_states_yelp}
\end{figure}
\end{document}